\def\year{2021}\relax
\documentclass[letterpaper]{article} % DO NOT CHANGE THIS
\usepackage{aaai21}  % DO NOT CHANGE THIS
\usepackage{times}  % DO NOT CHANGE THIS
\usepackage{helvet} % DO NOT CHANGE THIS
\usepackage{courier}  % DO NOT CHANGE THIS
\usepackage[hyphens]{url}  % DO NOT CHANGE THIS
\usepackage{graphicx} % DO NOT CHANGE THIS
\usepackage{natbib}  % DO NOT CHANGE THIS AND DO NOT ADD ANY OPTIONS TO IT
\usepackage{caption} % DO NOT CHANGE THIS AND DO NOT ADD ANY OPTIONS TO IT
\usepackage[T1]{fontenc}    % use 8-bit T1 fonts
\usepackage[caption=false]{subfig} 
\usepackage[usenames,dvipsnames]{xcolor}
\usepackage[utf8]{inputenc} % allow utf-8 input
\usepackage{algorithmic}
\usepackage{algorithm}
\usepackage{amsfonts}       % blackboard math symbols
\usepackage{amsmath}
\usepackage{array,multirow}
\usepackage{booktabs}       % professional-quality tables
\usepackage{enumitem}
\usepackage{microtype}      % microtypography
\usepackage{nicefrac}       % compact symbols for 1/2, etc.
\usepackage{pifont}% http://ctan.org/pkg/pifont
\usepackage{xspace}
\usepackage{dsfont}
\usepackage{ifpdf} 

\usepackage[switch]{lineno}

\newcommand{\modelscript}[2]{#2\xspace}  
\newcommand{\ux}{{\mathbf{x}}}
\newcommand{\ind}{\mathds{1}}

\begin{document}
\title{Active Bayesian Assessment of Black-Box Classifiers}
\author {
    % Authors
    Disi Ji\textsuperscript{\rm 1},  
    Robert L. Logan IV\textsuperscript{\rm 1}, 
    Padhraic Smyth\textsuperscript{\rm 1}, 
    Mark Steyvers\textsuperscript{\rm 2} \\
}
\affiliations {
    % Affiliations
    \textsuperscript{\rm 1}Department of Computer Science, University of California, Irvine \\
    \textsuperscript{\rm 2}Department of  Cognitive Sciences, University of California, Irvine \\
    disij@uci.edu, 
    rlogan@uci.edu, 
    smyth@ics.uci.edu,
    mark.steyvers@uci.edu
}
 
\maketitle

\begin{abstract}%
    Recent advances in machine learning have led to increased deployment of black-box classifiers across a wide variety of applications.
    In many such situations there is a critical need to both reliably assess the performance of these pre-trained models and to perform this assessment in a label-efficient manner (given that labels may be scarce and costly to collect). 
    In this paper, we introduce an active Bayesian  approach for assessment of classifier performance  to satisfy the desiderata of both reliability and label-efficiency.
    We begin by developing inference strategies to quantify uncertainty for common assessment metrics such as accuracy, misclassification cost, and   calibration error. We then propose a general framework for active Bayesian assessment using inferred uncertainty to guide  efficient selection of instances for labeling, enabling better performance assessment with fewer labels. We demonstrate  significant gains from  our proposed active Bayesian approach   via a series of systematic empirical experiments assessing the performance of modern neural classifiers (e.g., ResNet and BERT) on several standard image and text classification datasets.
\end{abstract}

\section{Introduction}
Complex machine learning models, particularly deep learning models, are now being applied to a variety of practical prediction problems ranging from diagnosis of medical images~\cite{kermany2018identifying} to autonomous driving~\cite{du2017fused}.
Many of these models are black boxes from the perspective of downstream users, for example, models developed remotely by commercial entities and hosted as a service in the cloud~\cite{yao2017complexity,sanyal2018tapas}.

For a variety of reasons (legal, economic, competitive), users of machine learning models increasingly may have no direct access to the detailed workings of the model, how the model was trained, or the training data.
In this context, careful attention needs to be paid to accurate, detailed and robust assessments of the quality of a model's predictions, such that the model can be held accountable by users. This is particularly true in the common scenario where the model is being deployed in an environment that does not necessarily distributionally match the data that the model was trained.

In real-world application scenarios,  labeled data for assessment is likely to be scarce and costly to collect, e.g., for a model being deployed in a diagnostic imaging context in a particular hospital where labeling requires expensive human expertise. Thus, it is important to be able to accurately assess the performance black-box classifiers in environments where there is limited availability of labeled data.
With this in mind we develop a framework for {\bf active Bayesian assessment} of black-box classifiers, using techniques from  active learning to efficiently select instances to label so that uncertainty of assessment can be reduced, and deficiencies of models such as low accuracy, high calibration error or high cost mistakes   can be quickly identified. 

The primary contributions of our paper are:
\begin{itemize}[nosep]
    \item We propose a general Bayesian framework to assess black-box classifiers, providing uncertainty quantification
    for quantities such as classwise accuracy, expected calibration error (ECE), confusion matrices, and performance comparisons across groups.
    \item We develop a methodology for  active Bayesian assessment for an array of fundamental tasks including (1) estimation of model performance; (2) identification of model deficiencies; (3) performance comparison between groups.
    \item We demonstrate that our proposed approaches need significantly fewer labels than baselines, via a series of experiments assessing the performance of modern neural classifiers (e.g., ResNet and BERT) on several standard image and text classification datasets.
\end{itemize}

\section{Notation} 

We consider classification problems with a feature vector $\ux$ and a class label $y \in \{1, \ldots, K\}$, e.g., classifying image pixels $\ux$ into one of $K$ classes.
We are interested in assessing the performance of a pretrained prediction model $M$ that makes predictions of $y$ given a feature vector $\ux$, where $M$ produces $K$ numerical scores per class in the form of a set of estimates of class-conditional probabilities ${p}_M(y = k | \ux), k = 1,\ldots,K$.
$\hat{y} = \arg \max_{k} {p}_M(y = k | \ux)$ is the classifier's label prediction for a particular input $\ux$. $s(\ux) = {p}_M(y = \hat{y}| \ux)$ is the {\bf score} of a model, as a function of $\ux$, i.e., the class probability that the model produces for its predicted class $\hat{y} \in \{1,\ldots,K\}$ given input $\ux$. This is also referred to as a model's {\bf confidence} in its prediction and can be viewed as a model's own estimate of its accuracy.
The model's scores in general need not be perfectly calibrated, i.e., they need not match the true probabilities $p(y= \hat{y} | \ux)$.
 
We focus in this paper on assessing the performance of a model\modelscript{ $M$}{} that is a black box, where we can observe the inputs $\ux$ and the outputs ${p}_M(y = k | \ux)$, but don't have any other information about \modelscript{the inner-workings of $M$}{its inner workings}.
Rather than learning a model itself we want to learn about the characteristics of a fixed model that is making predictions in a particular environment characterized by some underlying unknown distribution $p(\ux,y)$.

\section{Performance Assessment}
 
\paragraph{Performance Metrics and Tasks:}
We will use $\theta$ to indicate a {\bf performance metric} of interest, such as classification accuracy, true positive rate, expected cost, calibration error, etc. Our approach to assessment of a metric $\theta$ relies on the notion of disjoint {\bf groups} (or partitions) $g = 1,\ldots, G$ of the input space $\ux \in {\cal{R}}_g$, e.g., grouping by predicted class $\hat{y}$. 
For any particular instantiation of groups $g$ and metric $\theta$, there are three particular {\bf assessment tasks} we will focus on in this paper: (1) estimation, (2) identification, and (3) comparison.  

\vskip 0.2cm
\textbf{Estimation:}
Let $\theta_1,\ldots,\theta_G$ be the set of true (unknown) values for some metric $\theta$ for some grouping $g$. The goal of estimation is to assess the quality of a set of estimates $\hat{\theta}_1,\ldots,\hat{\theta}_G$ relative to their true values.
In  this paper we will focus on RMSE loss  $   \bigl( \sum_g p_g (\theta_g - \hat{\theta}_g)^2 \bigr)^{1/2}$ to measure estimation quality, where $p_g = p(\ux \in {\cal{R}}_g)$ is the marginal probability of a data point being in group $g$ (e.g., as estimated from unlabeled data) and  $ \hat{\theta}_g$ is a point estimate of the true $\theta_g$, e.g., a maximum a posteriori (MAP) estimate.

\vskip 0.2cm
\textbf{Identification:}
Here the goal is to identify extreme groups, e.g., $g^* = \arg \min_g \theta_g$, such as the predicted class with the lowest accuracy (or the highest cost, swapping max for min).  In general we will investigate  methods for finding the $m$ groups with highest or lowest values of a metric $\theta$. To compare the set of identified groups to the true set of $m$-best/worst groups, we can use (for example) ranking measures to evaluate and compare the quality of different identification methods.

\vskip 0.2cm
\textbf{Comparison:}
The goal here is to determine if the difference between two groups $g_1$ and $g_2$ is statistically significant, e.g., to assess if accuracy or calibration for one group is significantly better than another group for some black-box classifier. A measure of the quality of a particular assessment method in this context is to compare how often, across multiple datasets of fixed size, a method correctly identifies if a significant difference exists and, if so, its direction.

\paragraph{Groups:} There are multiple definitions of groups that are of interest in practice. One grouping of particular interest is where  groups correspond to a model's predicted classes, i.e., $g=k$, and the partition of the input space corresponds to the model's decision regions $\ux \in {\cal{R}}_k$, i.e., $\hat{y}(\ux)=k$. If $\theta$ refers to classification accuracy, then $\theta_k$ is the accuracy per predicted class. For prediction problems with costs, $\theta_k$ can be the expected cost per predicted class, and so on. 

Another grouping of interest for classification models is the set of groups $g$ that correspond to bins $b$ of a model's score, i.e., $s(\ux) \in \mbox{bin}_b, b = 1\ldots, B$, or equivalently $\ux \in {\cal{R}}_b$ where ${\cal{R}}_b$ is the region of the input space where model scores lie in score-bin $b$.
The score-bins can be defined in any standard way, e.g.,  equal width $1/B$ or equal weight $p(s(\ux)  \in \mbox{bin}_b) = 1/B$.
$\theta_b$ can be defined as the  accuracy per score-bin, which in turn can be related to the well-known expected calibration error (ECE, e.g.,~\citet{guo2017calibration}) as we will discuss in more detail later in the paper\footnote{We use ECE for illustration in our results since it is widely used in the recent classifier calibration literature, but other calibration metrics could also be used, e.g., see~\citet{kumar2019verified}.}.

In an algorithmic fairness context, for group fairness~\cite{hardt2016equality}, the groups $g$ can correspond to  categorical values of a protected attribute such as gender or race, and $\theta$ can be defined (for example) as accuracy or true positive rate per group. 

In the remainder of the paper, we focus on developing and evaluating the effectiveness of different methods for assessing groupwise metrics $\theta_g$.
In the two sections below, we first describe a flexible Bayesian strategy for assessing performance metrics $\theta$ in the context of the discussion above, then outline a general {\bf active assessment framework} that uses the Bayesian strategy to address the three assessment tasks (estimation, identification, and comparison) in a label-efficient manner.

\section{Bayesian Assessment}

\label{sec:assessment} 
We outline below a Bayesian approach to make posterior inferences about performance metrics given labeled data, where the posteriors on  $\theta$ can be used to support the assessment tasks of estimation, identification, and comparison. For simplicity we begin with the case where $\theta$ corresponds to accuracy and then extend to other metrics such as ECE. 

The accuracy  for a group $g$  can be treated as an unknown Bernoulli parameter $\theta_g$. Labeled observations  %$(\ux^{(i)},y^{(i)}),
$ (\ux_i, y_i), i = 1,\ldots, N_g$ are sampled randomly per group conditioned on $\ux_i \in \mathcal{R}_g$, leading to a binomial likelihood with binary accuracy outcomes $\ind(y_i, \hat{y}\modelscript{_M}{}_i) \in \{0,1\}$. The standard frequency-based estimate is
$\hat{\theta}_g = \frac{1}{N_g} \sum_{i=1}^{N_g} \ind(y_i, \hat{y}_i)$.

It is natural to consider Bayesian inference in this context, especially in situations where there is relatively little labeled data available per group. With a conjugate prior ${\theta}_g \sim \text{Beta}(\alpha_g,\beta_g)$ and a binomial likelihood on binary outcomes $\ind(y_i, \hat{y}_i)$, we can update the posterior distribution of $\theta_g$ in closed-form to $\text{Beta}(\alpha_g + r_g, \beta_g + N_g - r_g)$ where $r_g$ is the number of correct label predictions $\hat{y} = y$ by the model given  $N_g$ trials for group $g$. 
 
For metrics other than accuracy, we sketch the basic idea here for Bayesian inference for ECE and provide additional discussion in the Supplement.\footnote{Link to the Supplement: \url{https://arxiv.org/abs/2002.06532}}
ECE is defined as $\sum_{b=1}^B p_b |\theta_{b} - s_b|$ where $B$ is the number of bins (corresponding to groups $g$), $p_b$ is the probability of each bin $b$, and $\theta_b$ and $s_b$ are the accuracy and average confidence per bin respectively.  We can put  Beta priors on accuracies $\theta_{b}$, model  the likelihood of outcomes for each bin $b$  as binomial, resulting again in closed form Beta posteriors for accuracy per bin $b$. The posterior density for the marginal ECE itself is not available in closed form, but can easily be estimated by direct Monte Carlo simulation from the $B$ posteriors for the $B$ bins. We can also be Bayesian about ECE {\it per group}, $\mbox{ECE}_g$ (e.g., per class, with $g=k$),  in a similar manner by defining two levels of grouping, one at the class level and one at the bin level.% (see Supplement for details).

\begin{figure}[t]
    \centering
    \includegraphics[width=0.95\linewidth]{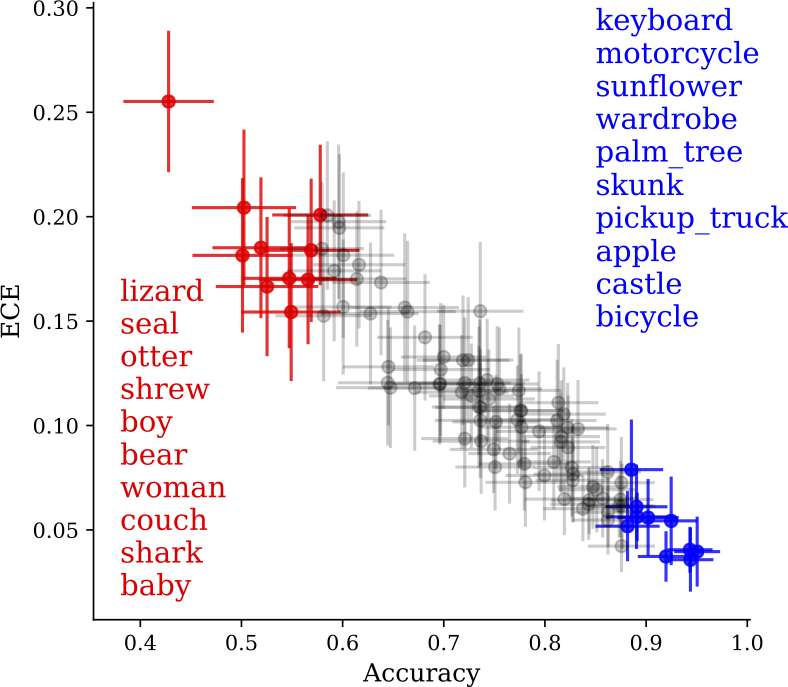}
    \caption{Scatter plot of estimated  accuracy and expected calibration error (ECE) per class of a ResNet-110 image classifier on the CIFAR-100 test set, using our Bayesian assessment framework, with posterior means and 95\% credible intervals per class. Red and blue for the top-10 least and most accurate classes, gray for the other classes.
    }
\label{fig:intro}
\end{figure}

\paragraph{Illustrative Example:}
To illustrate the general idea of Bayesian assessment, we train a standard ResNet-110 classifier on the CIFAR-100 data set and perform Bayesian inference of the accuracy and ECE of this model using the 10,000 labeled examples in the test set. The groups $g=k$ here correspond to the $K$ predicted classes by the model, $\hat{y}=k \in \{1,\ldots,K\}$. 
We use Beta priors with $\alpha_k=\beta_k=1, k=1,\ldots,K$ for classwise accuracy, and $\alpha_b = 2s_b, \beta_b = 2(1-s_b), b = 1, \ldots, K$ for binwise accuracy. 
The prior distributions reflect no particular prior belief about classwise accuracy and  a weak prior belief that the confidence of the classifier is calibrated across all predicted classes.

Figure~\ref{fig:intro} shows the resulting mean posterior estimates (MPEs) and 95\% credible intervals (CIs) for accuracy  and   ECE values  for each of the $K=100$ classes.  
The accuracies and ECE values of the model vary substantially across classes, and  classes with low  accuracy  tend to be less calibrated than those with higher accuracy. There is also considerable posterior uncertainty for these metrics even using the whole test set of CIFAR-100. For example, while there is   confidence that the least accurate class is ``lizard" (top left point), there is much less certainty about what class is the most accurate class (bottom right). 

It is straightforward to apply this type of Bayesian inference to other  metrics and to other assessment tasks, such as estimating a model's confusion matrix, ranking performance by groups  with uncertainty~\cite{marshall1998league}, analyzing significance of differences in performance across groups, and so on. For the CIFAR-100 dataset, based on the test data we can, for example,  say that with 96\% probability the ResNet-110 model is less accurate when predicting the superclass ``human'' than it is when predicting ``trees''; and that with 82\% probability, the accuracy of the model when it predicts  ``woman'' will be lower than the accuracy when it predicts ``man."
Additional details and examples   can be found in the Supplement.

\begin{table*}[t]
  \small
  \centering
    \begin{tabular}{llccccc}
    \toprule
    \multicolumn{1}{l}{} & Assessment Task & Prior: $p(\theta)$ &  Likelihood: $q_\theta(z|g)$ &  \multicolumn{2}{c}{Reward: $r(z|g)$}\\
    \midrule
    \multicolumn{1}{l}{Estimation} 
    &\multicolumn{1}{l}{Groupwise Accuracy}     
    &$\theta_g\sim\text{Beta}(\alpha_g, \beta_g)$
    &$z\sim\text{Bern}(\theta_g)$
    & \multicolumn{2}{c}{$p_g\cdot(\text{Var}(\hat\theta_g|\mathcal{L})-\text{Var}(\hat\theta_g|\{\mathcal{L}, z\}))$}\\
    
    \multicolumn{1}{l}{} 
    &\multicolumn{1}{l}{Confusion Matrix($g=k$)}       
    &$\theta_{\cdot k} \sim \text{Dirichlet}(\alpha_{\cdot k})$  
    &$z\sim\text{Multi}(\theta_k)$ 
    &\multicolumn{2}{c}{$p_k\cdot(\text{Var}(\hat\theta_k|\mathcal{L})-\text{Var}(\hat\theta_k|\{\mathcal{L}, z\}))$}\\
    
    \midrule
    
    \multicolumn{1}{l}{Identification} 
    &\multicolumn{1}{l}{Least Accurate Group}   
    &$\theta_g\sim\text{Beta}(\alpha_g, \beta_g)$
    &$z\sim\text{Bern}(\theta_g)$
    &\multicolumn{2}{c}{$-\widetilde\theta_g$}\\
    
    \multicolumn{1}{l}{} 
    &\multicolumn{1}{l}{Least Calibrated Group} 
    &$\theta_{gb}\sim\text{Beta}(\alpha_{gb}, \beta_{gb})$
    &$z\sim\text{Bern}(\theta_{gb})$
    &\multicolumn{2}{c}{$\sum_{b=1}^B p_{gb} \left|\widetilde\theta_{gb} - s_{gb}\right|$}\\
    
    \multicolumn{1}{l}{} 
    &\multicolumn{1}{l}{Most Costly Class($g=k$)}
    &$\theta_{\cdot k} \sim \text{Dirichlet}(\alpha_{\cdot k})$
    &$z\sim\text{Multi}(\theta_k)$ 
    &\multicolumn{2}{c}{$\sum_{j=1}^{K} c_{jk} \widetilde\theta_{jk}$}\\ 
    
    \midrule
    
    \multicolumn{1}{l}{Comparison} 
    &\multicolumn{1}{l}{Accuracy Comparison}    
    &$\theta_g\sim\text{Beta}(\alpha_g, \beta_g)$
    &$z\sim\text{Bern}(\theta_g)$
    &\multicolumn{2}{c}{$\lambda|\{\mathcal{L}, (g,z)\}$}\\
    
    \bottomrule
    
    \end{tabular}
    \caption{Thompson sampling configurations for different assessment tasks. 
    }
    \label{tab:thompson}
\end{table*}

\begin{figure}[t]
\centering
\begin{minipage}{\linewidth}
\begin{algorithm}[H]
  \caption{Thompson Sampling($p, q, r, M$)}
  \label{alg:thompson}
    \begin{algorithmic}[1]
      \STATE
      Initialize the priors on metrics $\{p_{0}(\theta_1),\ldots,p_{0}(\theta_g)\}$
      \FOR{$i=1,2,\cdots$}
            \STATE \# Sample parameters for the metrics $\theta$ 
            \STATE $\widetilde{\theta}_g \sim p_{i-1}(\theta_g), g = 1,\ldots, G$
            \STATE \# Select a group $g$ (or arm) by maximizing expected reward
            \STATE $\hat g \leftarrow \arg\max_{g} \mathbb{E}_{q_{\widetilde\theta}}[ r(z|g) ]$
            %=\{1,\cdots,G\}}
            \STATE \# Randomly select an input data point from  group $\hat g$ group and compute its predicted label
            \STATE $\ux_i \sim {\cal{R}}_{\hat g}$
            \STATE $\hat{y}_i(\ux_i) = \arg\max_k p_M(y=k | \ux_i)$
            \STATE \# Query to get a true label (pull arm $\hat g$)
            \STATE $z_i \leftarrow f(y_i, \hat{y}_i(\ux_i))$
            \STATE \# Update parameters of the $\hat g$th metric
            \STATE $p_{i}(\theta_{\hat g})\propto p_{i-1}(\theta_{\hat g})q(z_i|\theta_{\hat g})$
      \ENDFOR
    \end{algorithmic}
\end{algorithm}
\end{minipage}
\caption{An outline of the algorithm for active Bayesian assessment using multi-arm bandit Thompson sampling with arms corresponding to groups $g$.}
\label{fig:algorithm}
\end{figure}

\section{Active Bayesian Assessment}
\label{sec:active_assessment}
As described earlier, in practice we may wish to assess how well a blackbox classifier performs in a new environment where we have relatively little labeled data. For example, our  classifier may have been trained in the cloud by a commercial vendor and we wish to independently evaluate its accuracy (and other metrics) in a particular context.
Rather than relying on the availability of a large random sample of labeled instances for inference, as we did in the results in Figure~\ref{fig:intro}, we can improve data efficiency by leveraging the Bayesian approach to support {\it active assessment}, actively selecting examples $\ux$ for labeling in a data-efficient manner. Below we  develop active assessment approaches for the  three tasks of estimation, identification, and comparison.
Efficient active selection of examples for labeling is particularly relevant when we have a potentially large pool of unlabeled examples $\ux$ available, and have limited resources for labeling (e.g., a human labeler with limited time). 

The Bayesian framework described in the last section readily lends itself to be used in Bayesian active learning algorithms, by considering model assessment as a multi-armed bandit problem where each group $g$ corresponds to an arm or a bandit.
In Bayesian assessment in this context, there are two key building blocks: (i) the assessment algorithm's current beliefs (prior or posterior distribution) for the metric of interest $\theta_g\sim p(\theta_g)$, and (ii) a generative model (likelihood) of the labeling \textit{outcome} $z \sim q_\theta(z|g), \forall g$.

Instead of labeling randomly sampled data points from a pool of unlabeled data, we propose instead to actively select data points to be labeled by iterating between: 
(1) \textit{labeling}:  actively selecting a group $\hat g$ based on the assessment algorithm's current beliefs about $\theta_g$,  randomly selecting a data point $\ux_i \sim {\cal{R}}_{\hat g}$ and then querying its label; and
(2) \textit{assessment}:  updating our beliefs about the performance metrix $\theta_g$ given the outcome $z_i$.
This active selection approach requires defining a \textbf{reward  function} $r(z|g)$ for the revealed outcome $z$ for the $g$-th group.
For example, if the assessment task is to generate low variance estimates of groupwise accuracy, $r(z|g)$ can be formulated as the reduction in uncertainty about $\theta_g$, given an outcome $z$, to guide the labeling process. 

Our primary goal in this paper is to demonstrate the general utility of active assessment for performance assessment of black-box classifiers rather than comparing different active selection methods. With this in mind, we focus in particular on the framework of Thompson sampling~\cite{thompson1933likelihood,russo2018tutorial} since we found it to be more reliably efficient compared to other active selection methods such as epsilon-greedy and upper-confidence bound (UCB) approaches (additional discussion is provided in the Supplement).

Algorithm~\ref{alg:thompson} describes a general active assessment algorithm based on Thompson sampling.
At each step $i$, a set of values for metrics $\theta_g, 1 \ldots, G$, are sampled from the algorithm's current beliefs, i.e.,  $\widetilde\theta_g \sim p_{i-1}(\theta_g)$ (line 4).
As an example, when assessing groupwise accuracy, $p_{i-1}(\theta_g)$ represents the algorithm's belief (e.g., in the form of a posterior Beta distribution) about the accuracy for group $g$ given $i-1$ labeled examples observed so far. The sampling step is a key difference between Thompson sampling and alternatives that use a point estimate to represent current beliefs (such as greedy approaches).

Conditioned on the sampled $\widetilde\theta_g$ values, the algorithm then selects the group $\hat g$ that maximizes the expected reward $\hat g = \arg \max_g \mathbb{E}_{q_{\widetilde\theta_g}}[ r(z| g) ]$ (line 6)  where $r(z|g)$ is task-specific and where $z\sim q_{\widetilde\theta_{\hat{g}}}(z|{\hat{g}})$ is the likelihood for outcome $z$. The algorithm then draws an input data point $\ux_i$ randomly from ${\cal{R}}_{\hat g}$, and uses the model $M$ to generate a predicted label $\hat{y}_i$.  The Oracle is then queried (equivalent to ``pulling arm ${\hat g}$" in a bandit setting) to obtain a label outcome $z_i$ and the algorithm's belief is updated (line 13) to update the posterior for $\theta_{\hat g}$.  

Note that this proposed algorithm implicitly assumes that the $\theta_g$'s are independent (by modeling beliefs about $\theta_g$'s independently rather than jointly). In some situations there may be additional information across groups $g$ (e.g., hierarchical structure) that could be leveraged  (e.g., via contextual bandits) to improve  inference but we leave this for future work.

We next discuss how specific reward functions $r$ can be designed for different assessment tasks of interest, with a summary provided in  Table~\ref{tab:thompson}. 

\paragraph{Estimation:}
The MSE for estimation accuracy for $G$ groups  can be written in bias-variance form as $\sum_{g=1}^G p_g \bigl( \mbox{Bias}^2(\hat{\theta}_g)  + \mbox{Var}(\hat{\theta}_g) \bigr)$. Given a fixed labeling budget  the bias term can be assumed to be small relative to the variance  (e.g., see~\citet{sawade2010active}), by using relatively weak priors for example.
It is straightforward to show that to  minimize $\sum_{g=1}^G p_g  \mbox{Var}(\hat{\theta}_g)$  the optimal number of labels per group $g$ is proportional to $\sqrt{p_g \theta_g (1 - \theta_g)}$, i.e., sample more points from larger groups and from groups where $\theta_g$ is furthest from 0 or 1. While the group sizes $p_g$ can be easily estimated from  unlabeled data, the $\theta_g$'s are  unknown, so we can't compute the optimal sampling weights a priori. Active assessment in this context allows one  to   minimize MSE (or RMSE) in an adaptive sequential manner. In particular we can do this by defining a reward function 
$r(z|g) = p_g\cdot(\text{Var}(\hat\theta_g|\mathcal{L})-\text{Var}(\hat\theta_g|\{\mathcal{L}_, z\}))$, where $\mathcal{L}$ is the set of labeled data seen to date, with the goal of selecting examples for labeling  that minimize the overall posterior variance at each step.
For confusion matrices, a similar argument applies but with multinomial likelihoods and Dirichlet posteriors on vector-valued $\theta_g$'s per group (see Table 1).
Although we did not develop methods to directly estimate ECE in an active manner, we can nonetheless assess it by actively estimating bin-wise accuracy using the framework above.

\paragraph{Identification:}
To identify the best (or worst performing) group, $\hat g = \arg \max_g \theta_g$, we can define a reward function using the sampled metrics $\widetilde\theta_g$ for each group. 
For example, to identify the least accurate class, the expected reward of the $g$-th group is  $\mathbb{E}_{q_{\widetilde\theta}}[ r(z_i)|g ]
= q_{\widetilde\theta}(y=1)(-\widetilde\theta_g) + q_{\widetilde\theta}(y=0)(-\widetilde\theta_g) 
= -\widetilde\theta_g.$
Similarly, because the reward functions of other identification tasks (Table~\ref{tab:thompson}) are independent of the value of $y$, when the assessment tasks are to identify the group with the highest ECE or misclassification cost,   maximization of the reward function corresponds to selecting the group with the greatest sampled ECE or misclassification cost.

In addition we experimented with a modified version of Thompson sampling (TS) that is designed for best-arm identification, called top-two Thompson sampling (TTTS)~\cite{russo2016simple}, but found that   TTTS and TS gave very similar results---so we just focus on TS in the results presented in this paper.

To extend this approach to identification of   the {\it best-$m$ arms}, instead of selecting the arm with the greatest expected reward, we pull the top-$m$-ranked arms   at each step, i.e. we query the true labels of $m$  samples, one sample $\ux$ randomly drawn from each of the top $m$ ranked groups. This best-$m$ approach can be seen as an application of the general best-$m$ arms identification method proposed by~\citet{komiyama2015optimal} for the problem of extreme arm identification. They proposed this multiple-play Thompson sampling (MP-TS) algorithm as a multiple-play multi-armed bandit problem,  and proved that MP-TS has the optimal regret upper bound when the reward is binary. 

\paragraph{Comparison:}
For the task of comparing differences in a performance metric $\theta$ between two groups, an active assessment algorithm can learn about the accuracy of each group by sequentially allocating the labeling budget between them.
Consider two groups $g_1$ and $g_2$  with a true accuracy difference  
$\Delta = \theta_{g_1} - \theta_{g_2}$. Our approach uses the ``rope'' (region of practical equivalence) method of Bayesian  hypothesis  testing (e.g.,~\citet{benavoli2017time}) as follows.
The cumulative density in each of three regions $\mu = (P(\Delta < -\epsilon), P(-\epsilon\le\Delta\le\epsilon)$,$P(\Delta>\epsilon))$ represents the 
posterior probability that the accuracy of group $g_1$ is more than $\epsilon$ lower than the accuracy of $g_2$, that the two accuracies are ``practically equivalent,'' or that $g_1$'s accuracy is more than $\epsilon$ higher than that of $g_2$, where $\epsilon$ is user-specified. In our experiments  we use $\epsilon=0.05$ and the cumulative densities $\mu$ are estimated with 10,000 Monte Carlo samples.

The assessment task is to identify the region $\eta = \arg\max(\mu)$ in which $\Delta$ has the highest cumulative density, where $\lambda = \max(\mu) \in [0,1]$ represents the confidence of the assessment.
Using Thompson sampling to actively select labels from $g_1$ and $g_2$, at the $i$-th step, when we get a $z_i$ for a data point from the $ g$-th group we update the Beta posterior of $\theta_{g}$.
The resulting decrease in uncertainty about $\theta_{g}$  depends on the realization of the binary variable $z_i$ and the current distribution of $\theta_{ g}$. 
We use $\lambda$ to measure the amount of evidence we gathered from the labeled data from both of the groups. 
Then we can select the group in a greedy manner that has the greater expected increase $\mathbb{E}_{q_{\widetilde\theta}} [ \lambda | \{\mathcal{L}, (g,z) ] - \mathbb{E}_{q_{\widetilde\theta}} [\lambda | \mathcal{L}]$, which is equivalent to selecting the arm with the largest $\mathbb{E}_{q_{\widetilde\theta}} [\lambda | \{\mathcal{L}, (g, z) ]$.
This approach of \textit{maximal expected model change strategy} has also been used in prior work in active learning for other applications~\cite{freytag2014selecting,vezhnevets2012active}.

\section{Experimental Settings}
%\paragraph
\begin{table}[t]
\small
\centering
\begin{tabular}{rcccc}%c}
\toprule
                    &           & Test Set  & Number of   &Prediction\\
                    & Mode      & Size        & Classes   & Model  $M$
                    \\
\midrule
     CIFAR-100      & Image & 10K   & 100       & ResNet-110                \\
     ImageNet       & Image & 50K   & 1000      & ResNet-152                \\
     SVHN           & Image & 26K  & 10        & ResNet-152                \\
     20 News        & Text  & 7.5K  & 20        & BERT\textsubscript{BASE}  \\
     DBpedia        & Text  & 70K   & 14        & BERT\textsubscript{BASE}  \\
\bottomrule
\end{tabular}
\caption{Datasets and models used in experiments.}
\label{tab:datasets}
\end{table} 

\paragraph{Datasets and Prediction Models:}
In our experiments we use a number of well-known image and text classification datasets, for both image classification (\emph{CIFAR-100}~\cite{krizhevsky2009learning},
    \emph{SVHN}~\cite{netzer2011reading} and
    \emph{ImageNet}~\cite{russakovsky2015imagenet}) and  
text classification (\emph{20 Newsgroups}~\cite{lang1995newsweeder} and
    \emph{DBpedia}~\cite{zhang2015character}).

For models $M$ we use ResNet~\cite{he2016deep} to perform image classification and BERT~\cite{devlin2019bert} to perform text classification.
%(additional details in the Supplement)
Each model is trained on standard training sets used in the literature and assessment is performed on random samples from the corresponding test sets.
Table~\ref{tab:datasets} provides a summary of datasets, models, and test sizes.
In this paper we focus on assessment of deep neural network models in particular since they are of significant current interest in machine learning---however,  our approach is broadly applicable to classification models in general.

Unlabeled data points $\ux_i$ from the test set are assigned to groups (such as predicted classes or score-bins) by each prediction model. Values for $p_g$ (for use in active learning in reward functions and in evaluation of assessment methods) are estimated using the  model-based assignments of test datapoints to groups. Ground truth values for $\theta_g$ are defined using the full labeled test set for each dataset.

\paragraph{Priors:} 
We investigate both uninformative and informative priors to specify prior distributions over groupwise metrics. All of the priors we use are relatively weak in terms of prior strength, but as we will see in the next section, the informative priors can be very effective when there is little labeled data available. We set the prior strengths as $\alpha_g + \beta_g = N_0 = 2$ for Beta priors and $\sum \alpha_g = N_0 = 1$ for Dirichlet priors in all experiments, demonstrating the robustness of the settings across a wide variety of contexts. For groupwise accuracy, the informative Beta prior for each group is $\text{Beta}(N_0 s_g, N_0(1-s_g))$, where $s_g$ is the average model confidence (score) of all unlabeled test data points for group $g$.
The uninformative prior distribution is $\alpha = \beta = N_0/2$.

For confusion matrices, there are $\mathcal{O}(K^2)$ prior parameters in total for $K$ Dirichlet distributions, each  distribution parameterized by a $K$-dimensional vector $\alpha_j$. As an informative prior for a confusion matrix we use the model's own prediction scores on the unlabeled test data, $\alpha_{jk} \propto \Sigma_{\mathbf{\ux} \in \mathcal{R}_k} {p}_M(y = j | \ux)$.
The uninformative prior for a confusion matrix is set as $\alpha_{jk} = N_0/K, \forall j, k$. In the experiments in the next section we show that even though our models are not well-calibrated (as is well-known for deep models, e.g.,~\citet{guo2017calibration}; see also Figure 1), the model's own estimates of class-conditional probabilities nonetheless contain valuable information about confusion probabilities.

\section{Experimental Results}

\begin{table}[t] 
\small
% \fontsize{9}{11}\selectfont
\centering
%\ra{0.4}
%\resizebox{\columnwidth}{!}
 
\begin{tabular}{@{}rccccc@{}}
\toprule 
 & {N/K} & {N} & {    UPrior}& {    IPrior}& { IPrior+TS}\\ 
 &   &   & {     (baseline) }& {    (ours)}& { (ours)}\\ 
\midrule
{           CIFAR-100} & 2 & 200 & 30.7 &\textbf{15.0} &15.3 \\ 
{                    } & 5 & 500 & 20.5 &\textbf{13.6} &13.8 \\ 
{                    } & 10 & 1K & 13.3 &\textbf{10.9} &11.4 \\ 
\midrule
{            ImageNet} & 2 & 2K & 29.4 &13.2 &\textbf{13.2} \\ 
{                    } & 5 & 5K & 18.8 &12.1 &\textbf{11.6} \\ 
{                    } & 10 & 10K & 11.8 &9.5 &\textbf{9.4} \\ 
\midrule
{                SVHN} & 2 & 20 & 13.7 &5.1 &\textbf{3.4} \\ 
{                    } & 5 & 50 & 7.7 &5.1 &\textbf{3.4} \\ 
{                    } & 10 & 100 & 5.4 &4.7 &\textbf{3.1} \\ 
\midrule
{             20 News} & 2 & 40 & 23.9 &12.3 &\textbf{11.7} \\ 
{                    } & 5 & 100 & 15.3 &10.8 &\textbf{10.3} \\ 
{                    } & 10 & 200 & 10.4 &\textbf{8.7} &8.8 \\ 
\midrule
{             DBpedia} & 2 & 28 & 14.9 &2.0 &\textbf{1.5} \\ 
{                    } & 5 & 70 & 3.5 &2.3 &\textbf{1.2} \\ 
{                    } & 10 & 140 & 2.6 &2.1 &\textbf{1.1} \\ 
\bottomrule
\end{tabular}

\caption{RMSE of classwise accuracy  across 5 datasets. Each RMSE number is the mean across 1000 independent runs.%The strength of priors is 2.
}
\label{tab:classwise_acc_estimate}  
\end{table}

\begin{table}[t]
\small
\centering
\begin{tabular}{@{}rcccccc@{}}
\toprule 
& {N/K} & {N}& {    UPrior}& {    IPrior}& { IPrior+TS}\\ 
&   &   & {     (baseline) }& {    (ours)}& { (ours)}\\ 
\midrule
{           CIFAR-100} & 2 & 200 & 1.463 &0.077 &\textbf{0.025} \\ 
{                    } & 5 & 500 & 0.071 &0.012 &\textbf{0.004} \\ 
{                    } & 10 & 1K & 0.001 &0.002 &\textbf{0.001} \\ 
\midrule
{                SVHN} & 2 & 20 & 92.823 &0.100 &\textbf{0.045} \\ 
{                    } & 5 & 50 & 11.752 &0.022 &\textbf{0.010} \\ 
{                    } & 10 & 100 & 0.946 &0.005 &\textbf{0.002} \\ 
\midrule
{             20 News} & 2 & 40 & 3.405 &0.018 &\textbf{0.005} \\ 
{                    } & 5 & 100 & 0.188 &0.004 &\textbf{0.001} \\ 
{                    } & 10 & 200 & 0.011 &0.001 &\textbf{0.000} \\ 
\midrule
{             DBpedia} & 2 & 28 & 1307.572 &0.144 &\textbf{0.025} \\ 
{                    } & 5 & 70 & 33.617 &0.019 &\textbf{0.003} \\ 
{                    } & 10 & 140 & \textbf{0.000} &0.004 &0.001 \\ 
\bottomrule
\end{tabular}
\caption{
Scaled mean RMSE for confusion matrix estimation.
Same setup as Table~\ref{tab:classwise_acc_estimate}.
}
\label{tab:confusion_rmse}
\end{table}

We conduct a series of experiments across datasets, models, metrics, and assessment tasks, to systematically compare three different assessment methods: (1) non-active sampling with uninformative priors (UPrior), (2) non-active sampling with  informative priors (IPrior), and (3) active Thompson sampling (Figure~\ref{fig:algorithm}) with informative priors (IPrior+TS). Estimates of metrics (as used for example in computing RMSE or ECE)  correspond to mean posterior estimates $\hat{\theta}$ for each method.
Note that the UPrior method is equivalent to standard frequentist estimation with random sampling and weak additive smoothing. We use UPrior instead of a pure frequentist method to avoid numerical issues in very low data regimes.

Best-performing values that are statistically significant, across the three methods, are indicated in bold in our tables. Statistical significance between the best value and next best is determined by a Wilcoxon signed-rank test with p=0.05. Results are statistically significant in all rows in all tables, except for SVHN results in Table~\ref{tab:difference}. Code and scripts for all of our experiments are available at \url{https://github.com/disiji/active-assess}.

Our primary goal is to evaluate the effectiveness of active versus non-active assessment, with a secondary goal of evaluating the effect of informative versus non-informative priors.
As we will show below, our results clearly demonstrate that the Bayesian and active assessment frameworks are significantly more label-efficient and accurate, compared to the non-Bayesian non-active alternatives, across a wide array of assessment tasks.

\paragraph{Estimation of Accuracy, Calibration, and Confusion Matrices:}
We compare the estimation efficacy of each evaluated method as the labeling budget $N$ increases, for   classwise accuracy (Table~\ref{tab:classwise_acc_estimate}),  confusion matrices (Table~\ref{tab:confusion_rmse}), and ECE (Table~\ref{tab:ece_estimate}). All reported numbers are obtained by averaging across 1000 independent runs, where a run corresponds to a sequence of sampled $\ux_i$ values (and sampled $\theta_g$ values for the TS method).

Table~\ref{tab:classwise_acc_estimate} shows the mean RMSE of classwise accuracy for the 3 methods on the 5 datasets. The results demonstrate that informative priors and active sampling have significantly lower RMSE than the baseline, e.g., reducing RMSE by a factor of 2 or more in the low-data regime of $N/K=2$. Active sampling (IPrior+TS) improves on the IPrior method in 11 of the 15 results, but the gains are typically small. For other metrics and tasks below we see much greater gains from using active sampling.

Table~\ref{tab:confusion_rmse} reports the mean RMSE across runs of  estimates of confusion matrix entries for four datasets\footnote{ImageNet is omitted because 50K labeled samples is not sufficient to estimate a confusion matrix that contains ~1M parameters.}.
RMSE is defined here as 
$\text{RMSE} = \bigl(\sum_k p_k  \sum_j (\theta_{jk} - \hat{\theta}_{jk})^2\bigr)^{1/2}$ where $\theta_{jk}$ is the  probability that class $j$ is the true class when class $k$ is predicted.
To help with interpretation, we scale the  errors in the table by a constant $\theta_0$, defined as the RMSE of the confusion matrix estimated with scores from only unlabeled data, i.e. the estimate with IPrior when $N=0$.  
Numbers greater than 1  mean that the estimate is worse than using $\theta_0$ (with no labels). 

The results show that informed priors (IPrior and IPrior+TS)  often produce RMSE values that are orders of magnitude lower than using a simple uniform prior (UPrior). Thus, the model scores  on the unlabeled test set (used to construct the informative priors) are highly informative for confusion matrix entries, even though the models themselves are (for the most part) miscalibrated. We see in addition that active sampling (IPrior+TS) provides additional significant reductions in RMSE over the IPrior method with no active sampling.

For DBpedia, with a uniform prior and randomly selected examples, the scaled mean RMSE is 0.000. One plausible explanation is  that the accuracy of the classifier on DBpedia is 99\%, resulting in a confusion matrix that is highly diagonally dominant, and this ``simple structure'' results in an easy estimation problem once there are at least a few labeled examples.

\begin{table}[t]
\small
\centering
\begin{tabular}{@{}rccccc@{}}
\toprule 
 & {N/K} & {N}& {    UPrior}& {    IPrior}& { IPrior+TS}\\ 
  &   &   & {     (baseline) }& {    (ours)}& { (ours)}\\ 
\midrule
{           CIFAR-100} & 2 & 20 & 76.7 &\textbf{26.4} &28.7 \\ 
{                    } & 5 & 50 & 40.5 &\textbf{23.4} &26.7 \\ 
{                    } & 10 & 100 & 25.7 &\textbf{21.5 }&23.2 \\ 
\midrule
{            ImageNet} & 2 & 20 & 198.7 &51.8 &\textbf{36.4} \\ 
{                    } & 5 & 50 & 122.0 &55.3 &\textbf{29.6} \\ 
{                    } & 10 & 100 & 66.0 &40.8 &\textbf{22.1} \\ 
\midrule
{                SVHN} & 2 & 20 & 383.6 &86.2 &\textbf{49.7} \\ 
{                    } & 5 & 50 & 155.8 &93.1 &\textbf{44.2} \\ 
{                    } & 10 & 100 & 108.2 &80.6 &\textbf{36.6} \\ 
\midrule
{             20 News} & 2 & 20 & 54.0 &\textbf{39.7} &46.1 \\ 
{                    } & 5 & 50 & 32.8 &\textbf{28.9} &36.6 \\ 
{                    } & 10 & 100 & 24.7 &\textbf{22.3} &28.7 \\ 
\midrule
{             DBpedia} & 2 & 20 & 900.3 &118.0 &\textbf{93.1} \\ 
{                    } & 5 & 50 & 249.6 &130.5 &\textbf{74.5} \\ 
{                    } & 10 & 100 & 169.1 &125.9 &\textbf{60.9} \\ 
\bottomrule
\end{tabular}
\caption{Mean percentage estimation error of ECE with bins as groups.  Same setup as Table~\ref{tab:classwise_acc_estimate}.}
\label{tab:ece_estimate}
\end{table}

In our bin-wise accuracy experiments samples are grouped into 10 equal-sized bins according to their model scores. The active assessment framework allows us to estimate bin-wise accuracy with actively selected examples. 
By comparing the bin-wise accuracy estimates with bin-wise prediction confidence, we can then generate estimates of ECE to measure the amount of miscalibration of the classifier. 
We report error for overall ECE rather than error per score-bin since ECE is of more direct interest and more interpretable.
Table~\ref{tab:ece_estimate}  reports the average relative ECE estimation error, defined as
$ (100/R) \sum_{r=1}^R | \mbox{ECE}_N - \hat{\mbox{ECE}}_r | / \mbox{ECE}_N $ where $ \mbox{ECE}_N$ is the ECE measured on the full test set, and $\hat{\mbox{ECE}}_r$ is the esimated ECE (using MPE estimates of $\theta_b$'s), for a particular method on the $r$th run, $r=1,\ldots,R=1000$. 
Both the IPrior and IPrior+TS methods have significantly lower percentage error in general in their ECE estimates compared to the naive UPrior baseline, particularly on the three image  datasets (CIFAR-100, ImageNet, and SVHN). The bin-wise RMSE of the estimated $\theta_b$'s are reported in the Supplement and show similar gains for IPrior and IPrior+TS.

\paragraph{Identification of Extreme Classes:}
For our identification experiments, for a particular metric and choice of groups,  we conducted 1000 different sequential runs.   For each run, after each labeled sample, we rank the estimates  $\hat{\theta}_g$ obtained from each of the three methods, and compute the mean-reciprocal-rank (MRR) relative to the true top-$m$ ranked groups (as computed from the full test set).  The MRR of the predicted top-$m$ classes is defined as
$MRR = \frac{1}{m} \sum_{i=1}^{m} \frac{1}{\text{rank}_i}$
where $\text{rank}_i$ is the predicted rank of the $i$th best class.

Table~\ref{tab:best-m-acc} shows the mean percentage of labeled test set examples needed to correctly identify the target classes where ``identify" means the minimum number of labeled examples required so that the
% target classes are identified when 
MRR is greater than 0.99. For all 5 datasets the active method (IPrior+TS) clearly outperforms the non-active methods, with large gains in particular for cases where the number of classes $K$ is large (CIFAR-100 and Imagenet). Similar gains in identifying the least calibrated classes are reported in the Supplement.

\begin{table}[t]
\small
\centering
\begin{tabular}{rcccc}
\toprule 
    & Top-$m$ & UPrior  &IPrior &IPrior+TS \\ 
  &   & {     (baseline) }& {    (ours)}& { (ours)}\\ 
\midrule
     CIFAR-100  &1&81.1 &83.4 &\textbf{24.9}  \\
                &10&99.8 &99.8 &\textbf{55.1} \\ \midrule
      ImageNet  &1&96.9 &94.7 & \textbf{9.3}  \\
                &10&99.6 &98.5 &\textbf{17.1 }\\ \midrule
          SVHN  &1&90.5 &89.8 &\textbf{82.8}  \\
                &3&100.0 &100.0 &\textbf{96.0} \\ \midrule
       20 News  &1&53.9 &55.4 &\textbf{16.9 } \\
                &3&92.0 &92.5 &\textbf{42.5} \\ \midrule
       DBpedia  &1& 8.0 & \textbf{7.6} &11.6  \\
                &3&91.9 &90.2 &\textbf{57.1} \\ 
\bottomrule
\end{tabular}

\caption{Percentage of labeled samples needed to identify the least accurate top-1 and top-$m$ predicted classes across 5 datasets across 1000 runs.
}
\label{tab:best-m-acc}
\end{table} 

\begin{figure}[t]
    \centering 
    {\includegraphics[width=\linewidth,height=0.34\linewidth
    ]{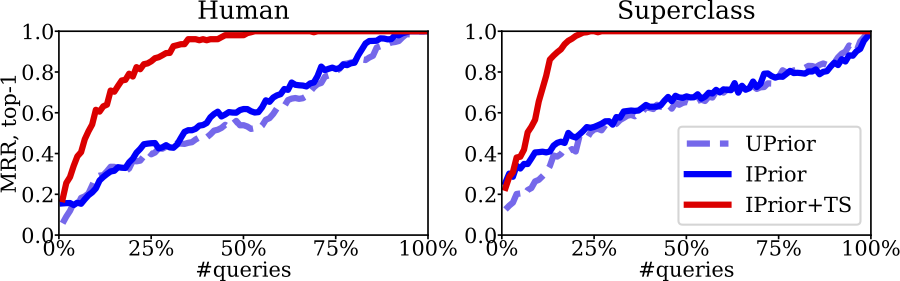}\\
    \vspace{0.3cm}
    \includegraphics[width=\linewidth,height=0.34\linewidth
    ]{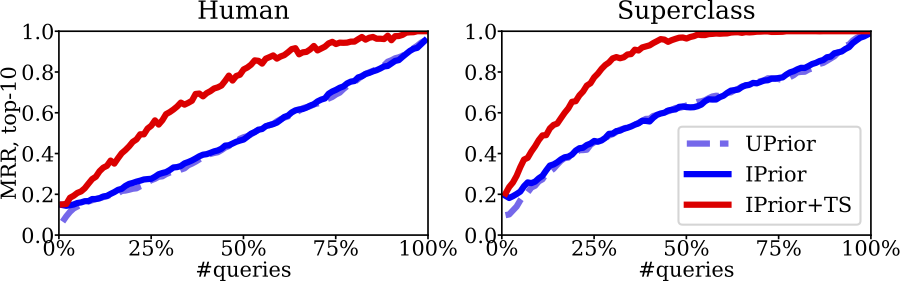}
    }
    \caption{
    MRR of 3 assessment methods for identifying the top 1 (top) and top 10 (bottom) highest-cost predicted classes, with 2 different cost matrices (right and left), averaged over 100 trials. See text for details.
    \label{fig:cost}
    }
\end{figure}

Figure~\ref{fig:cost} compares our 3 assessment methods for identifying the predicted classes with highest expected cost, using data from CIFAR-100, with two different (synthetic) cost matrices. In this plot the x-axis is the number of labels $L_x$ (queries) and the y-value is the average (over all runs) of the MRR conditioned on $L_x$ labels.
In the  left column (Human) the cost of misclassifying a person (e.g., predicting \textit{tree} when the true class is a \textit{woman}, etc.) is 10 times more expensive than other mistakes. In the right column, costs are 10 times higher if a prediction error is in a different superclass than the superclass of the true class (for the 20 superclasses in CIFAR-100).
The curves show the MRR as a function of the number of labels (on average, over 100 runs) for each of the 3 assessment methods. The active assessment method (IPrior+TS) is much more efficient at identifying the highest cost classes than the two non-active methods. 
The gains from active assessment are also robust to different settings of the relative costs of mistakes (details in the Supplement).

\begin{table}[t]
\small
\centering
\begin{tabular}{@{}rccc@{}}
\toprule 
& {    UPrior}& {    IPrior}& { IPrior+TS}\\ 
& (baseline) & (ours) & (ours) \\ \midrule
{CIFAR-100, Superclass} & 203.5 &129.0 &\textbf{121.9} \\ 
{                SVHN} & 391.1 &205.2 &172.0 \\ 
{             20 News} & 197.3 &157.4 &\textbf{136.1} \\ 
{             DBpedia} & 217.5 &4.3 &\textbf{2.8} \\ 
\bottomrule
\end{tabular}

\caption{Average number of labels across all pairs of classes required to estimate $\lambda$ for randomly selected pairs of predicted classes.
\label{tab:difference}}
\end{table}

\paragraph{Comparison of Groupwise Accuracy:}\label{section:active_difference}
For comparison experiments, Table~\ref{tab:difference} shows the results for the number of labeled data points required by each method  to reliably assess the accuracy difference of two predicted classes, averaged over independent runs for all pairwise combinations of classes. The labeling process terminates when the most probable region $\eta$ is identified correctly and the estimation error of the cumulative density $\lambda$ is within 5\% of its value on the full test set.
The results show that actively  allocating a labeling  budget and informative priors always improves label efficiency over uniform priors with no active assessment. 
In addition, active sampling (IPrior+TS) shows a systematic reduction of 5\% to 35\%  in the mean number of labels required across datasets,  over non-active sampling (IPrior).

%%% Related Work %%%
\section{Related Work}

\paragraph{Bayesian and Frequentist Classifier Assessment:} Prior work on Bayesian assessment of prediction performance, using Beta-Bernoulli models for example, has focused on specific aspects of performance modeling, such as estimating precision-recall performance~\cite{goutte2005probabilistic},  comparing classifiers~\cite{benavoli2017time}, or analyzing performance  of diagnostic tests~\cite{johnson2019gold}. ~\citet{welinder2013lazy} used a Bayesian approach to leverage a classifier's scores on unlabeled data for Bayesian evaluation of performance, and 
~\citet{ji2020can} also used Bayesian estimation with scores from unlabeled data to assess group fairness of blackbox classifiers in a label-efficient manner.
%and developed a hierarchical calibration model to leverage unlabeled data to improve label-efficiency.

Frequentist methods for label-efficient evaluation of classifier performance have included techniques such as importance sampling~\cite{sawade2010active} and stratified sampling~\cite{kumar2018classifier}, and low-variance sampling methods have been developed for evaluation of information retrieval systems~\cite{aslam2006statistical,yilmaz2006estimating,moffat2007strategic}. 

The framework we develop in this paper significantly generalizes these earlier contributions, by addressing a broader range of metrics and performance tasks within a single Bayesian assessment framework and by introducing the notion of  active assessment  for label-efficiency.

\paragraph{Active Assessment:}
While there is a large literature on active learning and multi-armed bandits (MAB) in general, e.g.,~\cite{settles2012active, russo2018tutorial},  our work is the first to our knowledge that applies ideas from Bayesian active learning to general classifier assessment, building on MAB-inspired, pool-based active learning algorithms for data selection.  ~\citet{nguyen2018active} developed  non-Bayesian active learning methods to select samples for estimating visual recognition performance of an algorithm on a fixed test set and similar ideas have been explored in the information retrieval literature~\cite{sabharwal2017good, li2017active,rahman2018efficient,voorhees2018building,rahman2019constructing}. This prior work is relevant to the ideas proposed and developed in this paper, but  narrower in scope in terms of performance metrics and tasks.

\section{Conclusions} 
In this paper we described a Bayesian framework for assessing performance metrics of black-box classifiers, developing inference procedures for an array of assessment tasks.
In particular, we proposed a new framework called {\it active assessment} for label-efficient assessment of classifier performance, and demonstrated significant performance improvements across five well-known datasets using this approach.

There are a number of interesting  directions for future work, such as Bayesian estimation of continuous functions related to accuracy and calibration, rather than using discrete groups, and Bayesian assessment in a sequential non-stationary context (e.g., with label  and/or covariate shift). The framework can also be extended to assess the same black-box model operating in multiple environments using a Bayesian hierarchical approach, or to comparatively assess multiple models operating in the same environment. 

\section*{Acknowledgements}
% This work was supported in part by the National Science Foundation [???]. The content is solely the responsibility of the authors and does not necessarily represent the official views of the National Science Foundation.
% This material is based upon work supported in part by the National Science Foundation under grants number 1900644 and 1927245, by the Defense Advanced Research Projects Agency (DARPA) under Contract No. HR001120C002, and by a Qualcomm Faculty Award (PS). This work was also partially funded by the Center for Statistics and Applications in Forensic Evidence (CSAFE) through Cooperative Agreement 70NANB20H019 between NIST and Iowa State University, which includes activities carried out at the University of California,  Irvine.
This material is based upon work supported in part by the National Science Foundation under grants number 1900644 and 1927245, by the Defense Advanced Research Projects Agency (DARPA) under Contract No. HR001120C002, by the Center for Statistics and Applications in Forensic Evidence (CSAFE) through Cooperative Agreement 70NANB20H019, and by a Qualcomm Faculty Award (PS).
%Carnegie Mellon University, Duke University,  University of Virginia, West Virginia University, University of Pennsylvania, Swarthmore College and University of Nebraska, Lincoln.
%Additional funding related to this work include research funding from NIH, NASA, PCORI, SAP and eBay, honoraria for lectures for General Motors, consulting income from Toshiba, and internships at Google and Facebook.

\clearpage

\section*{Broader Impact}
Machine learning classifiers are currently widely used to make predictions and decisions across a wide range of applications in society: education admissions, health insurance, medical diagnosis, court decisions, marketing, face recognition, and more---and this trend is likely to continue to grow. When these systems are deployed in real-world environments it will become increasingly important for users to have the ability to perform reliable, accurate, and independent evaluation of the performance characteristics of these systems and to do this in a manner which is efficient in terms of the need for labeled data. y

Our paper addresses this problem directly, providing a general-purpose and transparent framework for label-efficient performance evaluations of black-box classifier systems. The probabilistic (Bayesian) aspect of our approach provides users with the ability to understand how much they can trust performance numbers given a fixed data budget for evaluation.  For example, a hospital system or a university might wish to evaluate multiple different performance characteristics of pre-trained classification models in the specific context of the population of patients or students in their institution. The methods we propose have the potential to contribute to increased societal trust in AI systems that are based on machine learning classification models.

\section*{Supplemental Material for Section 4: Bayesian Assessment}

\subsection*{Bayesian Metrics}
\paragraph{Bayesian Estimates of Reliability Diagrams}

One particular application of Bayesian groupwise accuracy estimation is to \textbf{reliability diagrams}. Reliability diagrams are a widely used tool for visually diagnosing model calibration~\cite{degroot1983comparison,niculescu2005predicting}.
These diagrams plot the empirical sample accuracy $A_M(\ux)$ of a model $M$ as a function of the model's confidence scores $s_M(\ux)$.
If the model is perfectly calibrated, then $A_M(\ux) = s_M(\ux)$ and the diagram consists of the identity function on the diagonal. % (apart from measurement noise due to empirical sampling).
Deviations  from the diagonal reflect miscalibration of the model.
In particular if the curve lies below the diagonal with $A_M(\ux) < s_M(\ux)$ then the model $M$ is overconfident (e.g., see \citet{guo2017calibration}).
For a particular value $s_M(\ux) = s \in [0,1]$ along the x-axis, the corresponding y value is defined as: $\mathbb{E}_{\ux | s_M(\ux)=s} [A_M(\ux)]$.

To address data sparsity, scores are often aggregated into bins. 
We use equal-width bins here, denoting the $b$-th bin or region as $\mathcal{R}_b = \{\ux | s_M(\ux) \in [(b-1)/B, b/B)\}$, where $b = 1,\ldots, B$ ($B=10$ is often used in practice).
The unknown accuracy of the model per bin is $\theta_b$, which can be viewed as a marginal accuracy over the region ${\mathcal{R}_b}$ in the input space corresponding to $s(\ux) \in \mathcal{R}_b$, i.e., 
$$\theta_b = \int_{{\mathcal{R}_b}} p( y =\hat{y}_M  | \ux) p(\ux | \ux \in {\mathcal{R}_b}) d\ux.$$
As described in the main paper, we can put Beta priors on each $\theta_b$ and define a binomial likelihood on outcomes within each bin (i.e., whether a model's predictions are correct or not on each example in a bin), resulting in posterior Beta densities for each $\theta_b$.

\begin{figure*}[ht]
    \centering
    \includegraphics[width=\linewidth]{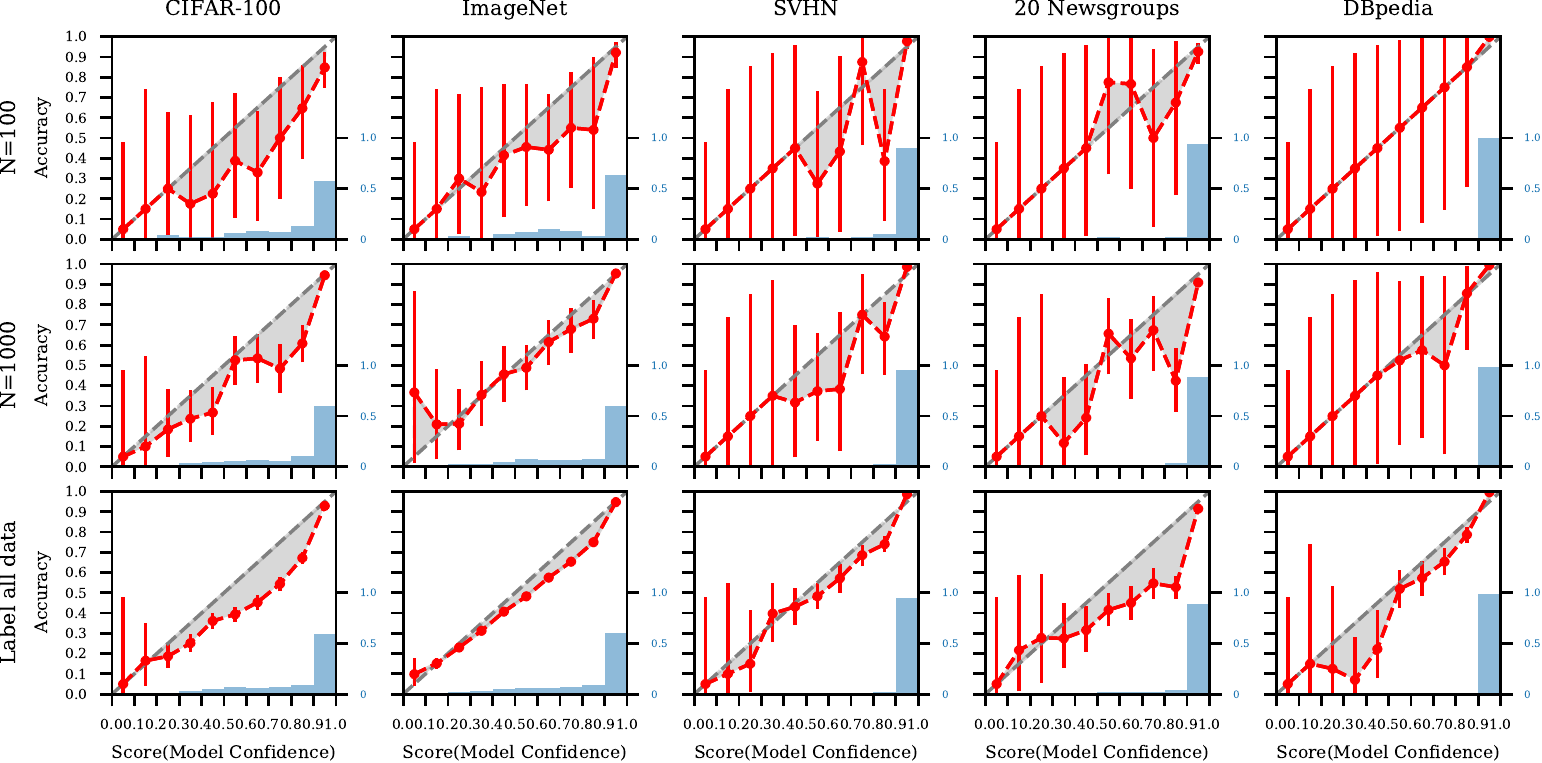} 
    \caption{Bayesian reliability diagrams for five datasets (columns) estimated using varying amounts of  test data (rows).
    The red solid circles plot the posterior mean for $\theta_j$ under the Bayesian model.
    Red bars display 95\% %posterior
    credible intervals.
    Shaded gray areas indicate the estimated magnitudes of the calibration errors, relative to the Bayesian estimates.
    The blue histogram shows the distribution of the scores for $N$ randomly drawn samples.
    }
    \label{fig:reliability}
\end{figure*}

In Figure~\ref{fig:reliability} we show the Bayesian reliability diagrams for the five datasets discussed in the main paper. %, based on  different amounts of labeled 
The columns indicate different datasets and the rows indicate how much data (from the test set) was used to estimate the reliability diagram.
Based on the full set of test examples (row 3), the posterior means and the posterior 95\% credible intervals are generally below the diagonal, i.e., we can infer with high confidence that the models are miscalibrated (and overconfident, to varying degrees) across all five datasets.  
For some bins where the scores are less than 0.5, the credible intervals are wide due to little data, and there is not enough information to determine with high confidence if the corresponding models are calibrated or not in these regions. 
With only $N = 100$ examples (row 1), the posterior uncertainty captured by the 95\% credible intervals indicates that there is not yet enough information to determine whether the models are miscalibrated.
With $N = 1000$ examples (row 2) there is enough information to reliably infer that the CIFAR-100 model is overconfident in all bins for scores above 0.3.
For the remaining datasets the credible intervals are generally wide enough to include 0.5 for most bins, meaning that we do not have enough data to make reliable inferences about calibration, i.e., the possibility that the models are well-calibrated cannot be ruled out without acquiring more data.

\paragraph{Bayesian Estimation of Calibration Performance (ECE)}
\begin{figure*}[t]
    \centering
    \includegraphics[width=\linewidth]{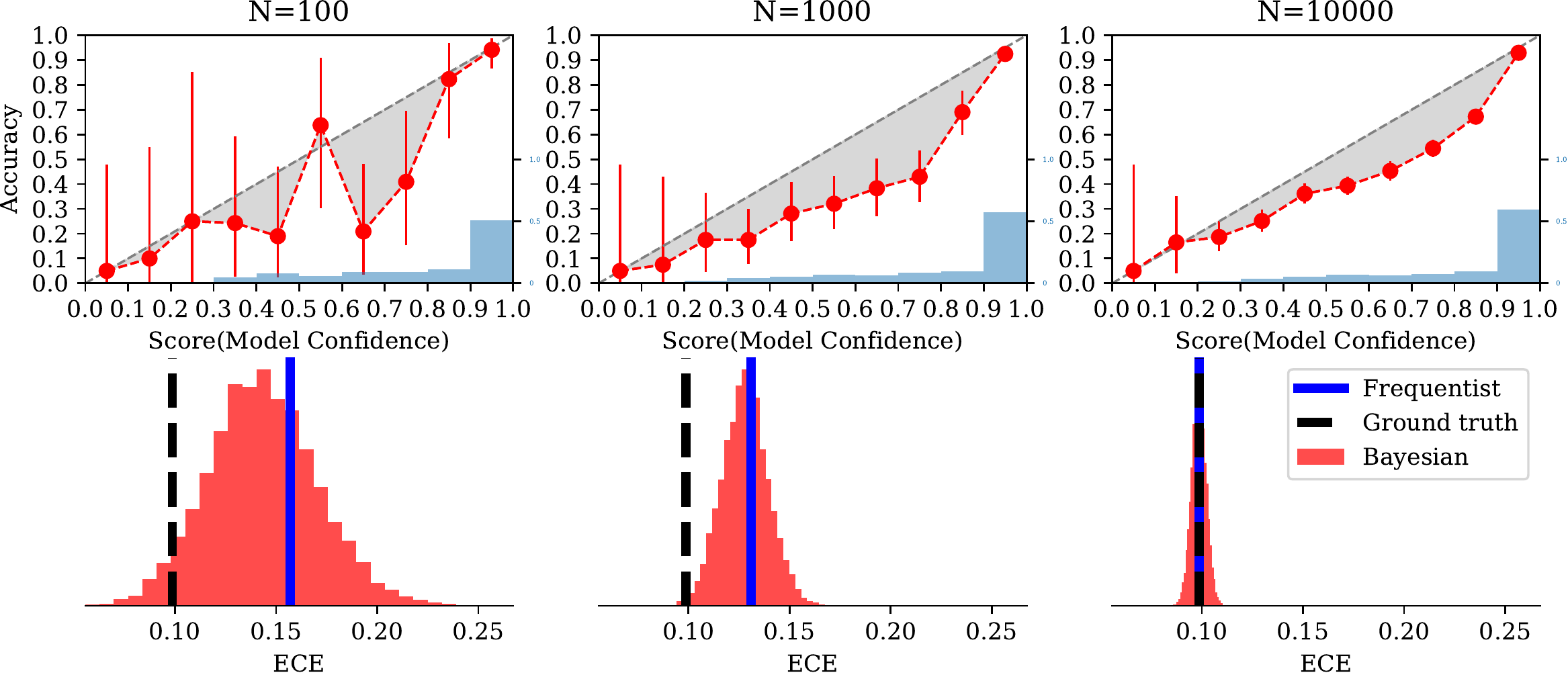}
    \caption{Bayesian reliability diagrams (top) and posterior densities for ECE (bottom) for CIFAR-100 as the amount of data used for estimation increases. Vertical lines in the right plots depict the ground truth ECE (black, evaluated with all available assessment data) and frequentist estimates (red).}
\label{fig:ece_posterior}
\vspace{1.0 cm}
    \centering
    \includegraphics[width=\linewidth]{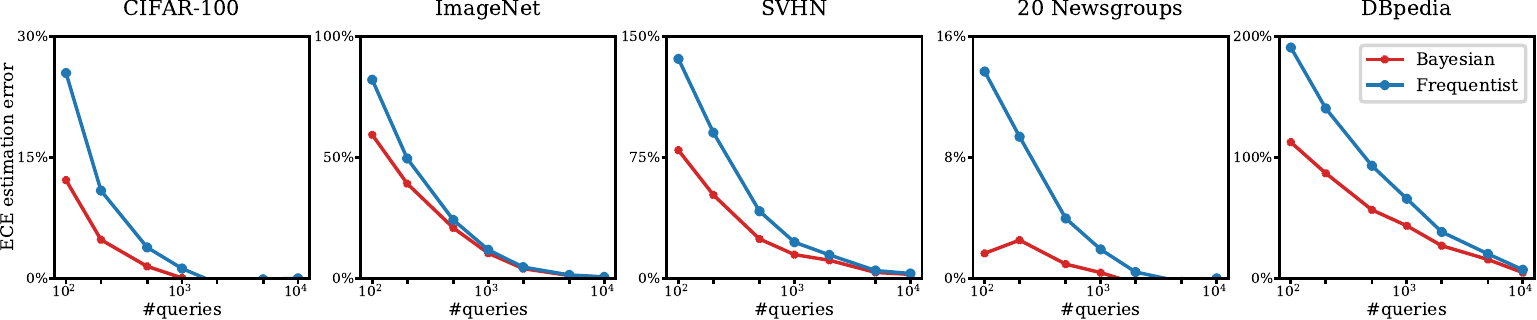}
    \caption{Percentage error in estimating expected calibration error (ECE) as a function of dataset size, for Bayesian (red) and frequentist (blue) estimators, across five datasets.}
\label{fig:reliability_comparison}
\vspace{1.0 cm}
 \centering
    \includegraphics[width=\linewidth]{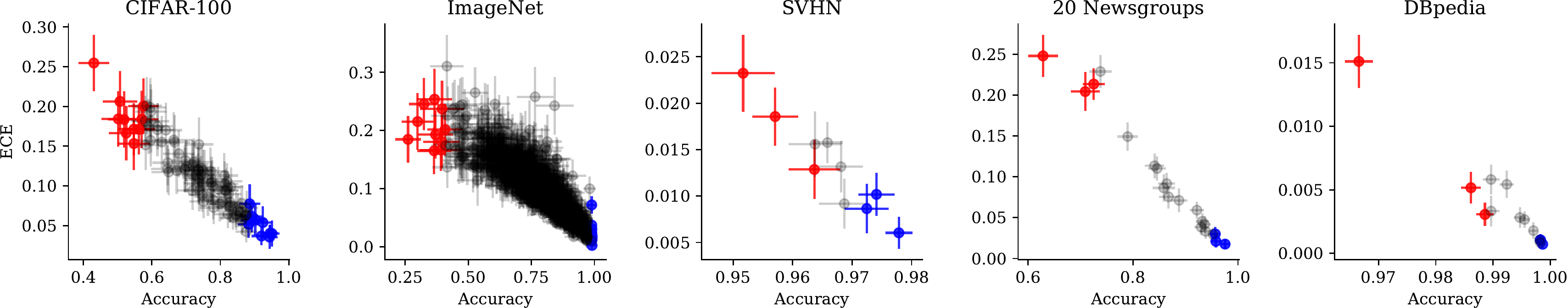}
\caption{
Scatter plots of classwise accuracy and ECE for the five datasets in the main paper. 
Each marker represents posterior means and 95\% credible intervals of posterior accuracy and ECE for each predicted class. Markers in red and blue represent the top-$m$ least and most accurate predicted classes, markers in gray represent the other classes, with $m = 10$ for CIFAR-100 and ImageNet, and $m = 3$ for the other  datasets.
}
\label{fig:scatter}

\end{figure*}
As shown in Figure~\ref{fig:reliability}, ``self-confident'' estimates provided by machine learning predictors can often be quite unreliable and miscalibrated \citep{zadrozny2002transforming,kull2017beta,snoek2019can}. In particular, complex models such as deep networks with high-dimensional inputs (e.g., images and text) can be significantly overconfident in practice~\citep{gal2016dropout,guo2017calibration,lakshminarayanan2017simple}.

We can assess calibration-related metrics for a classifier in a Bayesian fashion using any of well-known various calibration metrics, defined as the discrepancy between model scores and accuracy~\citep{kumar2019verified,Nixon_2019_CVPR_Workshops}. Here we focus on expected calibration error (ECE) given that it is among the widely-used calibration metrics in the machine learning literature (e.g., \citet{guo2017calibration,snoek2019can}). 

As discussed in the previous subsection, using the standard ECE binning procedure, the marginal ECE is defined as a weighted average of the absolute distance between the binwise accuracy $\theta_b$ and the average score $s_b$ per bin:
\begin{equation}
\mbox{ECE} = \sum_{b=1}^B p_b |\theta_b - s_b|
\label{equ:ece}
\end{equation}
where $p_b$ is the probability of a score lying in bin $b$ (which can be estimated using available unlabeled data). As shown in Figure~\ref{fig:ece_posterior}, the weights of these bins tend to be quite skewed in practice for deep neural networks models like ResNet-110. 

While the posterior of ECE is not available in closed form,  Monte Carlo samples for it are straightforward to obtain, by drawing random samples from the posterior Beta distributions on the $\theta_b$'s and deterministically computing ECE with Equation~\ref{equ:ece}.

Figure~\ref{fig:ece_posterior} shows the results of Bayesian estimation of a reliability diagram (top row) and the resulting posterior estimate of ECE (bottom row) for the CIFAR-100 dataset with three different values of $N$. The third column for $N=10000$ corresponds to using all of the data in the test set. The other 2 columns  of plots correspond to particular random samples of size $N=100$ and $N=1000$. The ECE value computed using all the test data ($N=10000$) is referred to as ground truth in all plots, ``Bayesian" refers to the methodology described in the paragraphs above, and ``frequentist" refers to the standard frequentist estimate of  ECE.

The bottom row of Figure~\ref{fig:ece_posterior} plots summaries of empirical samples (in red) from the posterior density of ECE as the amount of  data increases.  As $N$ increases the
posterior density for ECE converges to ground truth, and the uncertainty about ECE decreases. 
When the number of samples is small ($N=100$), the Bayesian posterior for ECE puts non-negative probability mass on the ground truth marginal ECE, while the frequentist method significantly overestimates ECE without any notion of uncertainty. 

Figure~\ref{fig:reliability_comparison}  shows the percentage error in estimating ground truth ECE, for Bayesian mean posterior estimates (MPE) and frequentist estimates of ECE, as a function of the number of labeled data points (``queries'') across the five datasets in the paper.
The percentage is computed relative to the ground truth marginal $\text{ECE} = \text{ECE}^*$, computed with the whole test set  as before. 
The MPE is computed with Monte Carlo samples from the posterior distribution (histograms of such samples are shown in Figure~\ref{fig:ece_posterior}). 
At each step, we randomly draw and label $N$ queries from the pool of unlabeled data, and compute both a Bayesian and frequentist estimate of marginal calibration error using these labeled data points. We run the simulation 100 times, and report the average $\text{ECE}_N$ over the $N$ samples. Figure~\ref{fig:reliability_comparison} plots $(\text{ECE}_N - \text{ECE}^*)/\text{ECE}^*$ as a percentage. The Bayesian method consistently has lower ECE estimation error, especially when the number of queries is small. 

\paragraph{Bayesian Estimation of ECE per Class.}
Similar to accuracy, we can also model \emph{classwise ECE}, 
$$\text{ECE}_k = \sum_{b=1}^{B} p_{b,k} | \theta_{b,k} - s_b |$$ 
by modifying the model described above to use regions $\mathcal{R}_{b,k} = \{ \ux | \hat{y}=k, s(\ux) \in \mathcal{R}_b \}$ that partition the input space by predicted class in addition to partitioning by the model score. This follows the same procedure as for ``total ECE" in the previous subsection except that the data is now partitioned by predicted class $k=1,\ldots,K$ and a posterior density on $\text{ECE}_k$ for each class  is obtained.

In the main paper, we showed a scatter plot of classwise accuracy and classwise ECE assessed with our proposed Bayesian method for CIFAR-100. 
In Figure~\ref{fig:scatter} we show scatter plots for all five datasets used in the paper. The assessment shows that model accuracy and calibration vary substantially across classes. For CIFAR-100, ImageNet and 20 Newsgroups, the variance of classwise accuracy and ECE among all predicted class is considerably greater than the variance for the two other datasets. Figure~\ref{fig:scatter} also illustrates that there is significant negative correlation between classwise accuracy and ECE  across all 5 datasets, i.e. classes with low classwise accuracy also tend to be less calibrated.

\paragraph{Bayesian Estimation of Confusion Matrices}
\label{sec:bayesian-confusion}
Conditioned on a predicted class $\hat{y}$, the true class label $y$ has a categorical distribution $\theta_{jk} = p(y = j| \hat{y} = k)$. We will refer to $\theta_{jk}$ as confusion probabilities. % since they resemble the elements of a confusion matrix.
In a manner similar to using a beta-binomial distribution to model accuracy, we can model these confusion probabilities using a Dirichlet-multinomial distribution:
\begin{equation}
\theta_{\cdot k} \sim \text{Dirichlet}(\alpha_{\cdot k})
\end{equation}
There are $\mathcal{O}(K^2)$ parameters in total in $K$ Dirichlet distributions, each of which is parameterized with a $K$ dimensional vector $\alpha_j$. 

\paragraph{Bayesian Misclassification Costs}
Accuracy assessment can be viewed as implicitly assigning a binary cost to model mistakes, i.e. a cost of 1 to incorrect predictions and a cost of 0 to correct predictions. In this sense, identifying the predicted class with lowest accuracy is equivalent to identifying the class with greatest expected cost. However, in real world applications, costs of different types of mistakes can vary drastically. For example, in autonomous driving applications, misclassifying a pedestrian as a crosswalk can have much more severe consequences than other misclassifications.

To deal with such situations, we extend our approach to incorporate an arbitrary cost matrix $\mathbf{C} = [ c_{jk} ]$, where $c_{jk}$ is the cost of predicting class $\hat{y} = k$ for a data point whose true class is $y = j$. The {\bf classwise expected cost} for predicted class $k$ is given by:
\begin{equation}
C^M_{\mathcal{R}_k} 
    = \mathbb{E}_{p(\ux,y | \ux \in \mathcal{R}_k)} [c_{jk}\ind(y = j)] 
    = \sum_{j=1}^{K} c_{jk} \theta_{jk}.
\label{equ:cost}
\end{equation}

The posterior of $C^M_{\mathcal{R}_k}$ is not available in closed form but Monte Carlo samples are straightforward to obtain, by randomly sampling $\theta_{\cdot k} \sim \text{Dirichlet}(\alpha_{\cdot k})$ and computing $C^M_{\mathcal{R}_k}$ deterministically with the sampled $\theta_{\cdot k}$ and the predefined cost matrix $\mathbf{C}$.

\paragraph{Bayesian Estimation of Accuracy Differences}
\label{sec:bayesian-difference}

\begin{figure*}[h]
    \centering
    \includegraphics[width=0.60\linewidth]{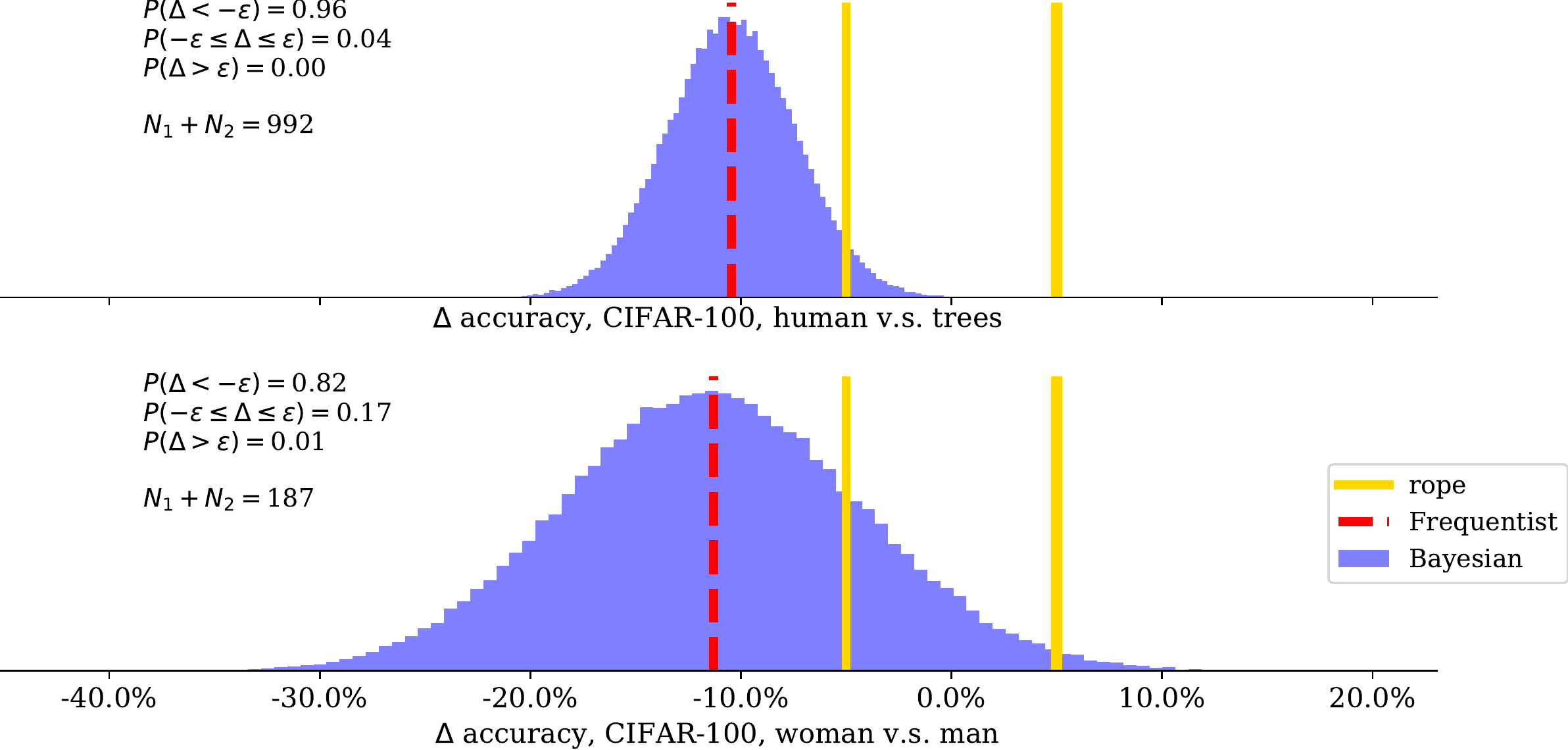}
    \caption{
    Density plot for the differences of accuracy between two superclasses/classes of CIFAR-100. Region of practical equivalence is [-0.05, 0.05]. Vertical solid lines in gold plots the region of practical equivalence, vertical red dashed line  plots the frequentist estimate of accuracy difference. Left: the two groups are the predicted superclasses ``human'' and ``trees'' in CIFAR100. Right: the two groups are the predicted classes ``woman'' and ``man'' in CIFAR-100. }
    \label{fig:rope}
\end{figure*}

Bayesian estimation of group differences allows us to compare the performance between two groups with uncertainty. For example, with a prior $\text{Beta}(1,1)$ and the full test set of CIFAR-100, the posterior distribution of groupwise accuracies of ResNet-110 on superclass ``human'' and ``trees'' are $\theta_{g_1}\sim \text{Beta}(280,203)$ and $\theta_{g_2}\sim \text{Beta}(351, 162)$ respectively. The total amount of labeled data for the two superclasses are 481 and 511. 
The difference in accuracy between superclass ``human'' and superclass ``trees'' is defined as $\Delta = \theta_{g_1} - \theta_{g_2}$. With random samples from the posterior distributions of $\theta_{g_1}$ and $\theta_{g_2}$, we can simulate the posterior distribution of $\Delta$ and compute its cumulative density in each of three regions $\mu = (P(\Delta < -\epsilon), P(-\epsilon\le\Delta\le\epsilon)$,$P(\Delta>\epsilon))$.
In Figure~\ref{fig:rope}, when $\epsilon=0.05$, ResNet-110 is less accurate on the predicted superclass ``human'' than on ``trees'' with 96\% probability.
Similarly with 82\% probability, the accuracy of ResNet-110 on ``woman'' is lower than ``man''.  
Although the point estimates of both performance differences have values that are approximately 10\%, the assessment of ``human'' versus ``tree'' is more certain because more samples are labeled.

\subsection*{Bayesian Assessment: Inferring Statistics of Interest via Monte Carlo Sampling}
An additional benefit of the Bayesian framework is that we can draw samples from the posterior to infer other statistics of interest. 
Here we illustrate this method with two examples.

\begin{figure*}[h]
 \centering
    \subfloat[]{\includegraphics[width=0.5\linewidth]{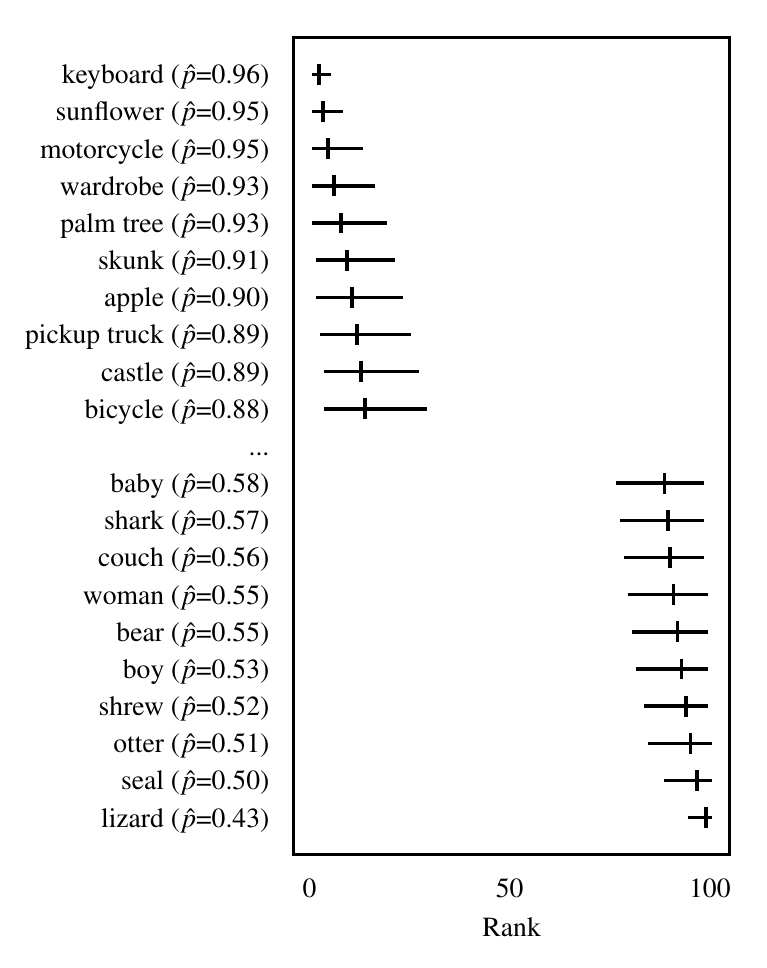}
    \label{fig:cifar100_rank}}
    \subfloat[]{\includegraphics[width=0.38\linewidth]{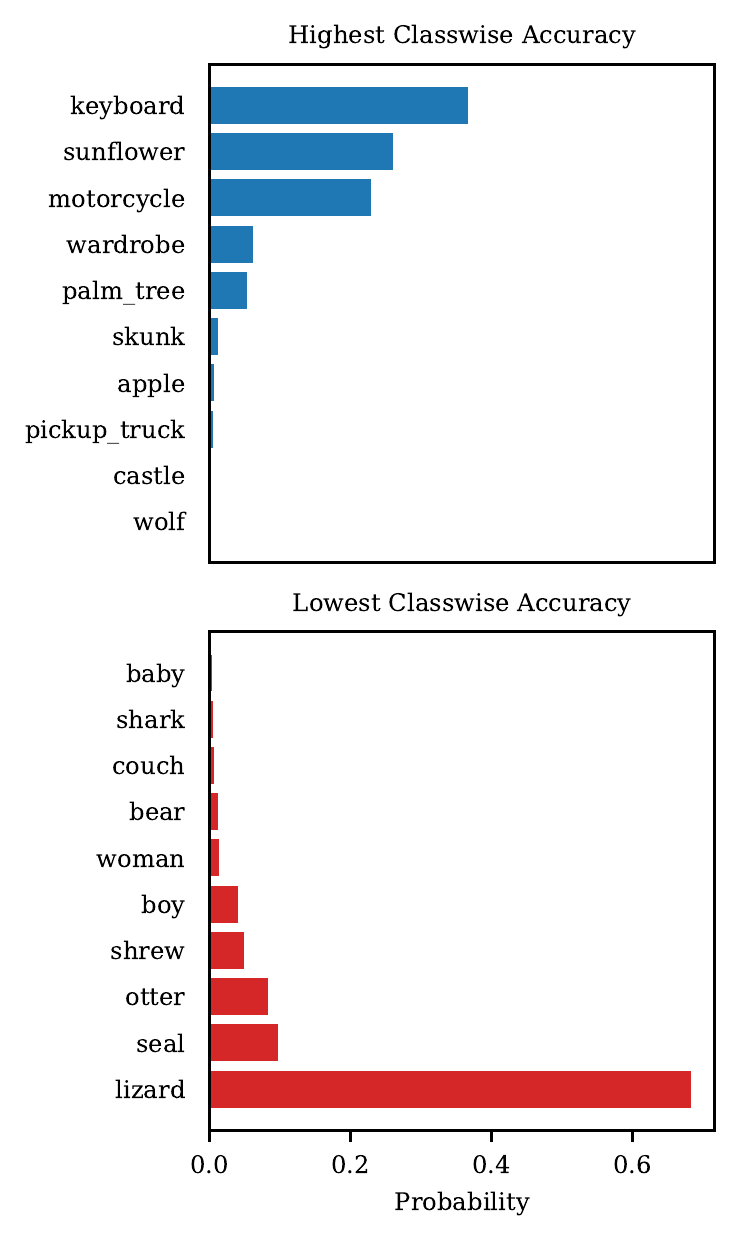}
    \label{fig:cifar100_top_bottom}}
    \caption{(a) MCMC-based ranking of accuracy across predicted classes for CIFAR-100 (where 1 corresponds to the class with the highest accuracy.
    (b) Posterior probabilities of the most and least accurate predictions on CIFAR-100. The class with the highest classwise accuracy is somewhat uncertain, while the class with the lowest classwise accuracy is very likely \textit{lizard}.}
\end{figure*}

\paragraph{Bayesian Ranking via Monte Carlo Sampling}
We can infer the Bayesian ranking of classes in terms of classwise accuracy or expected calibration error (ECE), by drawing samples from the posterior distributions (e.g., see \citet{marshall1998league}).
For instance, we can estimate the ranking of classwise accuracy of a model for CIFAR-100, by sampling $\theta_{k}$'s (from their respective posterior Beta densities) for each of the classes and then computing the rank of each class using the sampled accuracy.
We run this experiment 10,000 times and then for each class we can empirically estimate the distribution of its ranking.
The MPE and 95\% credible interval of ranking per predicted class for top 10 and bottom 10 are provided in Figure~\ref{fig:cifar100_rank} for CIFAR-100.

\paragraph{Posterior probabilities of the most and least accurate predictions}
We can estimate the probability that a particular class such as \emph{lizard} is the least accurate predicted class of CIFAR-100 by sampling $\theta_{k^*}$'s (from their respective posterior Beta densities) for each of the classes and then measuring whether $\theta_{lizard}$ is the minimum of the sampled values.
Running this experiment 10,000 times and then averaging the results, we determine that there is a 68\% chance that lizard is the least accurate class predicted by ResNet-110 on CIFAR-100.
The posterior probabilities for other classes are provided in Figure~\ref{fig:cifar100_top_bottom}, along with results for estimating which class has the highest classwise accuracy.

\clearpage
\section*{Supplemental Material for Section 5: Active Assessment}

\subsection*{Different Multi-Armed Bandit Algorithms for Best-Arm(s) Identification}
Below we provide brief descriptions for Thompson Sampling(TS)
and the different variants of multi-armed bandit algorithms for best arm identification and top-$m$ arm identification problems that we investigated in this paper, including Top-Two Thompson Sampling(TTTS)~\cite{russo2016simple} and multiple-play Thompson sampling(MP-TS)\citep{komiyama2015optimal}.

\paragraph{Best Arm Identification}
\begin{itemize}
    \item \textbf{Thompson sampling (TS)} is a widely used method for online learning of multi-armed bandit problems ~\citep{thompson1933likelihood,russo2018tutorial}. The algorithm samples actions according to the posterior probability that they are optimal. Algorithm 1 describes the sampling process for identifying the least accurate predicted class with TS.
    
    \item \textbf{Top-two Thompson sampling (TTTS)} is a modified version of TS that is tailored for best arm identification, and has some theoretical advantages over TS. Compared to TS, this algorithm adds a re-sampling process to encourage more exploration.\\
    Algorithm 2 describes the sampling process for identifying the least accurate predicted class with TTTS. The re-sampling process of TTTS is described in lines 10 to 24. At each step, with probability $1-\beta$ the algorithm selects the class $I$ which has the lowest sampled accuracy; in order to encourage more exploration, with probability $\beta$ the algorithm re-samples until a different class $J\neq I$ has the lowest sampled accuracy. 
    $\beta$ is a tuning parameter. 
    When $\beta=0$, there is no re-sampling in TTTS and it reduces to TS.
\end{itemize}
Figure~\ref{fig:ttts_compare} compares TS and TTTS for identifying the least accurate class for CIFAR-100. The results show that two methods are equally efficient across 5 datasets. 
For TTTS, we set the probability for re-sampling to $\beta=0.5$ as recommended in~\cite{russo2016simple}.

\paragraph{Top-$m$ Arms Identification}
\begin{itemize}
    \item \textbf{Multiple-play Thompson sampling (MP-TS)} is an extension of TS to multiple-play multi-armed bandit problems and it has a theoretical optimal regret guarantee with binary rewards. 
    Algorithm 3 is the sampling process to identify the least accurate $m$ arms with MP-TS, where $m$ is the number of the best arms to identify. 
    At each step, $m$ classes with the lowest sampled accuracies are selected, as describe in lines 10 to 20. 
    When $m=1$, MP-TS is equivalent to TS. 
\end{itemize}
In our experiments, we use TS for best arm identification and MP-TS for top-$m$ arms identification, and refer to both of the methods as TS in the main paper for simplicity.

\begin{figure}[h]
\centering
\begin{minipage}{\linewidth}
    \begin{algorithm}[H]
      \caption{Thompson Sampling (TS) Strategy}
      \label{alg:ts}
        \begin{algorithmic}[1]
        	\STATE {\bfseries Input:} prior hyperparameters $\alpha$, $\beta$
        	\STATE initialize $n_{k,0}=n_{k,1}=0$ for $k=1$ {\bfseries to} $K$
        	\REPEAT
        	\STATE \# Sample accuracy for each predicted class
        	\FOR{$k=1$ {\bfseries to} $K$}
                \STATE $\widetilde{\theta}_{k} \sim \text{Beta}(\alpha + n_{k,0}, \beta + n_{k,1})$
        	\ENDFOR
        	\STATE \# Select a class $k$ with the lowest sampled accuracy
        	\STATE ${\hat k} = \arg \min_k \widetilde{\theta}_{1:K}$
        	\STATE \# Randomly select an input data point from the ${\hat k}$-th class and compute its predicted label
            \STATE $\ux_i \sim {\cal{R}}_{\hat k}$
            \STATE $\hat{y}_i = \arg\max_k p_M(y=k | \ux_i)$
            \STATE \# Update parameters of the $\hat k$-th metric
        	\IF {$\hat{y}_i  = \hat k$}
                \STATE $n_{{\hat k},0} \leftarrow n_{{\hat k},0} + 1$
        	\ELSE
                \STATE $n_{{\hat k},1} \leftarrow n_{{\hat k},1} + 1$
            \ENDIF
        	\UNTIL{all data labeled}
        \end{algorithmic}
    \end{algorithm}
\end{minipage}
\caption{Thompson Sampling (TS) for identifying the least accurate class.}
\label{fig:ts}
\end{figure}

\begin{figure*}[ht]
    \centering
    \includegraphics[width=\linewidth]{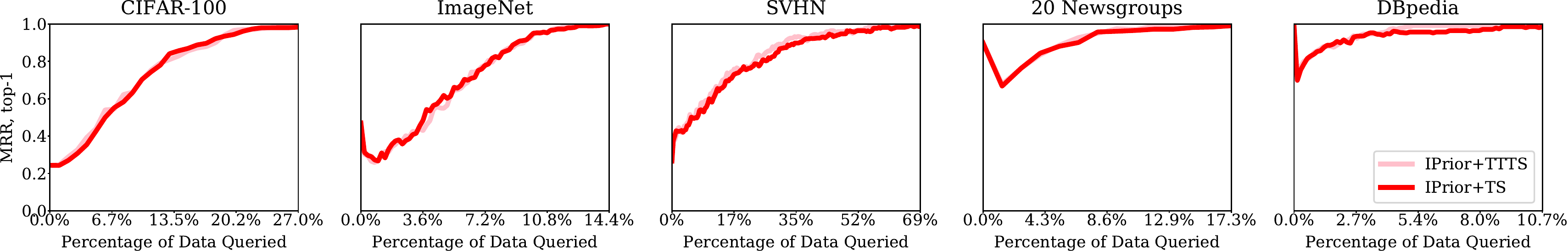}\\
    \caption{Mean reciprocal rank (MRR) of the class with the estimated lowest classwise accuracy with the strength of the  prior set as $\alpha+\beta = 2$, comparing TS and TTTS, across five datasets.
    The y-axis is the average MRR over 1000 runs for the percentage of queries, relative to the full test set, as indicated on the x-axis.  TS and TTTS have almost identical performance across the five datasets, as indicated by the graphs for each being on top of each other.
    }
    \label{fig:ttts_compare}
\end{figure*}

\begin{figure}[t]
\centering
\begin{minipage}{\linewidth}
    \begin{algorithm}[H]
      \caption{Top Two Thompson Sampling (TTTS) Strategy}
      \label{alg:ttts}
        \begin{algorithmic}[1]
        	\STATE {\bfseries Input:} prior hyperparameters $\alpha$, $\beta$
        	\STATE initialize $n_{k,0}=n_{k,1}=0$ for $k=1$ {\bfseries to} $K$
        	\REPEAT
        	\STATE \# Sample accuracy for each predicted class
        	\FOR{$k=1$ {\bfseries to} $K$}
                \STATE $\widetilde{\theta}_{k} \sim \text{Beta}(\alpha + n_{k,0}, \beta + n_{k,1})$
        	\ENDFOR
        	\STATE \# Select a class $k$ with the lowest sampled accuracy
        	\STATE $I = \arg \min_k \widetilde{\theta}_{1:K}$
        	\STATE \# Decide whether to re-sample
        	\STATE $B \sim \text{Bernoulli}(\beta$)
        	\IF {$B = 1$}
        	    \STATE \# If not re-sample, select $I$
        	    \STATE $\hat k = I$
        	\ELSE
        	    \STATE \# If re-sample, keep sampling until a different arm $J$ is selected
        	    \REPEAT
                	\FOR{$k=1$ {\bfseries to} $K$}
                        \STATE $\widetilde{\theta}_{k} \sim \text{Beta}(\alpha + n_{k,0}, \beta + n_{k,1})$
                	\ENDFOR
                	\STATE $J = \arg \min_k \widetilde{\theta}_{1:K}$
                \UNTIL{$J \neq I$}
                \STATE $\hat k = J$
        	\ENDIF
        	\STATE \# Randomly select an input data point from the ${\hat k}$-th class and compute its predicted label
            \STATE $\ux_i \sim {\cal{R}}_{\hat k}$
            \STATE $\hat{y}_i = \arg\max_k p_M(y=k | \ux_i)$
            \STATE \# Update parameters of the $\hat k$-th metric
        	\IF {$\hat{y}_i  = \hat k$}
                \STATE $n_{{\hat k},0} \leftarrow n_{{\hat k},0} + 1$
        	\ELSE
                \STATE $n_{{\hat k},1} \leftarrow n_{{\hat k},1} + 1$
            \ENDIF
        	\UNTIL{all data labeled}
        \end{algorithmic}
    \end{algorithm}
\end{minipage}
\caption{Top Two Thompson Sampling (TTTS) for identifying the least accurate class.}
\label{fig:ttts}
\end{figure}

\begin{figure}[t]
\centering
\begin{minipage}{\linewidth}
\begin{algorithm}[H]
  \caption{Multiple-play Thompson sampling (MP-TS) Strategy}
  \label{alg:mp-ts}
    \begin{algorithmic}[1]
    	\STATE {\bfseries Input:} prior hyperparameters $\alpha$, $\beta$
    	\STATE initialize $n_{k,0}=n_{k,1}=0$ for $k=1$ {\bfseries to} $K$
    	\REPEAT
    	\STATE \# Sample accuracy for each predicted class
    	\FOR{$k=1$ {\bfseries to} $K$}
            \STATE $\widetilde{\theta}_{k} \sim \text{Beta}(\alpha + n_{k,0}, \beta + n_{k,1})$
    	\ENDFOR
    	\STATE \# Select a set of $m$ classes with the lowest sampled accuracies
        \STATE $I^{*} = \text{top-m arms ranked by } \widetilde{\theta}_{k}$.
        \FOR{$\hat k \in I^{*}$}
        	\STATE \# Randomly select an input data point from the ${\hat k}$-th class and compute its predicted label
            \STATE $\ux_i \sim {\cal{R}}_{\hat k}$
            \STATE $\hat{y}_i = \arg\max_k p_M(y=k | \ux_i)$
            \STATE \# Update parameters of the $\hat k$-th metric
        	\IF {$\hat{y}_i  = \hat k$}
                \STATE $n_{{\hat k},0} \leftarrow n_{{\hat k},0} + 1$
        	\ELSE
                \STATE $n_{{\hat k},1} \leftarrow n_{{\hat k},1} + 1$
            \ENDIF
        \ENDFOR
    	\UNTIL{all data labeled}
    \end{algorithmic}
\end{algorithm}
\end{minipage}
\caption{Multiple-play Thompson Sampling (MP-TS) for identifying the least accurate $m$ classes.}
\label{fig:mp-ts}
\end{figure}

\clearpage
\newpage
\section*{Supplemental Material for Section 6: Experiment Settings}

\subsection*{Prediction models} 
For image classification we use ResNet \cite{he2016deep} architectures with either 110 layers (CIFAR-100) or 152 layers (SVHN and ImageNet). For ImageNet we use the pretrained model provided by PyTorch, and for CIFAR-100 and SVHN we use the pretrained model checkpoints provided at: \url{https://github.com/bearpaw/pytorch-classification}. For text classification tasks we use fine-tuned BERT\textsubscript{BASE} \cite{devlin2019bert} models.
Each model was trained on the standard training set used in the literature and assessment was performed on the test sets. 
Ground truth values for the assessment metrics were computed using the full labeled test set of each dataset.

\subsection*{Evaluation}
\begin{itemize}
    \item \textbf{Estimation}: we use RMSE of the estimated $\hat\theta$ relative to the true $\theta^*$ (as computed from the full test set) to measure the estimation error. 
    For Bayesian methods, $\hat\theta$ is the maximum posterior estimation(MPE) of $\theta$'s posterior distribution. For frequentist methods, $\hat\theta$ is the corresponding point estimation.\\
    The estimation error of groupwise accuracy and confusion matrix are defined as $\text{RMSE} = (\sum_g p_g(\hat\theta_g - \theta_g^*)^2)^{\frac{1}{2}}$ and $\text{RMSE} = (\sum_k p_k (\sum_j (\hat\theta_{jk} - \theta_{jk}^*)^2)^{\frac{1}{2}}$ respectively. 
    
    \item \textbf{Identification}: we compute the mean-reciprocal-rank (MRR) relative to the true top-$m$ ranked groups.  The MRR of the top-$m$ classes is defined as $MRR = \frac{1}{m} \sum_{i=1}^{m} \frac{1}{\text{rank}_i}$ where $\text{rank}_i$ is the predicted rank of the $i$th best class. Following standard practice, other classes in the best-$m$ are ignored when computing rank so that $MRR = 1$ if the predicted top-$m$ classes match ground truth. We set $m = 10$ for CIFAR-100 and ImageNet, and $m = 3$ for the other datasets.
    
    \item \textbf{Comparison}: we compare the results of rope assessment ($\eta, \lambda$) with the ground truth values ($\eta^*, \lambda^*$). The assessment is considered as a success if (1) the direction of difference is correctly identified $\eta = \eta ^*$ and (2) the estimation error of cumulative density is sufficiently small $|\lambda - \lambda^*| / \lambda^* < 0.05$.
\end{itemize}
In all experiments in our paper, unless stated otherwise, we report the aggregated performance averaged over 1000 independent runs.

\subsection*{Reproducibility}
We provide code to reproduce our results reported in the paper and in the Appendices. Our datasets and code are publicly available at: \url{https://github.com/disiji/active-assess}. The random seeds we used to generate the reported results are provided in the code. 

The memory complexity of the non-optimized implementation of Algorithm~\ref{alg:ts} and ~\ref{alg:mp-ts} is $\mathcal{O}(N+K)$, where $N$ is the number of data points and $K$ is the number of groups.
Overall the sampling methods we developed are computationally efficient. For example, for estimating groupwise accuracy of ResNet-110 on CIFAR-100, one run takes less than 10 seconds.  
All our experiments are conducted on Intel i9-7900X (3.3Ghz, 10 cores) with 32 GB of RAM.

Settings for hyperparameter and priors are discussed in Section ``Experimental Settings'' in the paper. 
We set the prior strengths as $\alpha_g + \beta_g = N_0 = 2$ for Beta priors and $\sum \alpha_g = N_0 = 1$ for Dirichlet priors in all experiments, unless otherwise stated, demonstrating the robustness of the settings across a wide variety of contexts.
In ``Experimental Results: Sensitivity Analysis for Hyperparameters'' of this Appendix  we provide additional sensitivity analysis.

\subsection*{Cost Matrices}

\begin{figure}[h]
    \centering 
    \includegraphics[width=\linewidth]{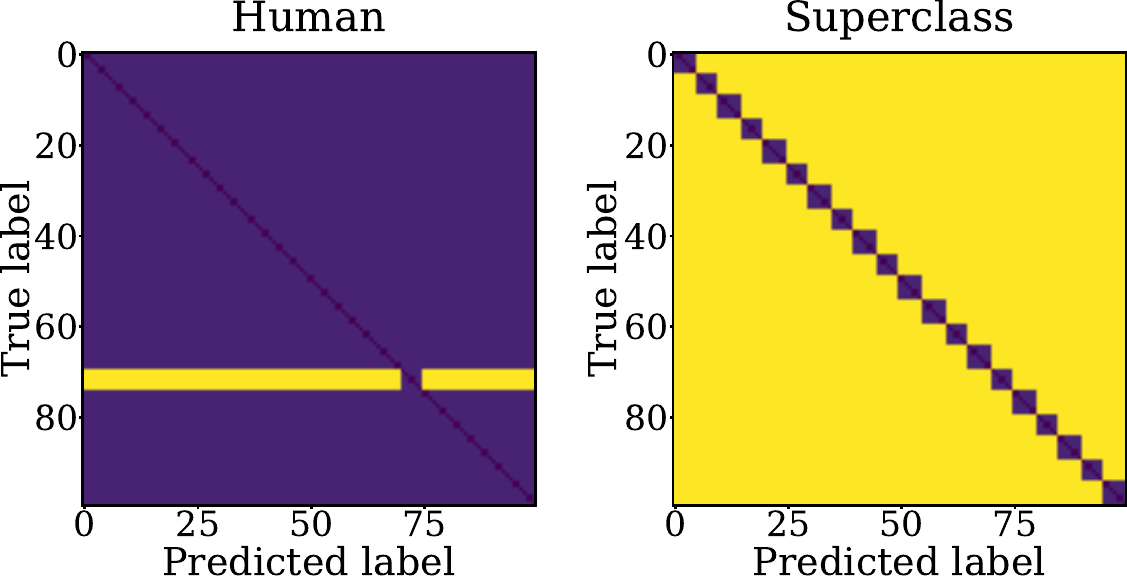}
    \caption{Cost matrices used in our experiments. (left): human,
    (right): superclass.
    }
    \label{fig:cost_matrix}
\end{figure}

To assess misclassification cost of the models, we experimented with 2 different cost matrices on the CIFAR-100 dataset:
\begin{itemize}
    \item \textbf{Human}: the cost of misclassifying a person (e.g., predicting \textit{tree} when the true class is a \textit{woman}, \textit{boy} etc.) is more expensive than other mistakes.
    \item \textbf{Superclass}: the cost of confusing a class with another superclass (e.g., a \textit{vehicle} with a \textit{fish}) is more expensive than the cost of mistaking labels within the same superclass (e.g., confusing \textit{shark} with \textit{trout}).
\end{itemize}
We set the cost of expensive mistakes to be 10x the cost of other mistakes.
In Figure~\ref{fig:cost_matrix}, we plot the two cost matrices.

\clearpage
\newpage

\section*{Supplemental Material for Section 7: Experimental Results}
\subsection*{RMSE of Binwise Accuracy Estimates}
In the main paper we report the estimation error results for overall ECE---here we also report results for error per score-bin. Table~\ref{tab:binwise_acc_estimate} provides the RMSE results for binwise accuracy estimates. The results demonstrate that both informative priors and active sampling significantly reduce RMSE relative to the baseline for all datasets and all $N$ values.

\begin{table}[h]
\centering{
\caption{RMSE of \textbf{binwise} accuracy estimates obtained with UPrior, IPrior and IPrior+TS across 5 datasets over 1000 independent runs. The strength of priors is 2.}
\resizebox{\columnwidth}{!}{%
\begin{tabular}{@{}cccccc@{}}
\toprule 
 & {N/K} & {N}& {    UPrior}& {    IPrior}& { IPrior+TS}\\ 
 &   &   & {     (baseline) }& {    (our work)}& { (our work)}\\ 
\midrule
{           CIFAR-100} & 2 & 20 & 23.2 &12.8 &\textbf{11.9} \\ 
{                    } & 5 & 50 & 15.7 &11.3 &\textbf{10.0} \\ 
{                    } & 10 & 100 & 11.0 &9.3 &\textbf{8.1} \\ 
\midrule
{             DBpedia} & 2 & 20 & 7.6 &2.3 &\textbf{1.4} \\ 
{                    } & 5 & 50 & 3.6 &2.5 &\textbf{1.2} \\ 
{                    } & 10 & 100 & 2.4 &2.1 &\textbf{0.9} \\ 
\midrule
{       20 Newsgroups} & 2 & 20 & 19.6 &11.5 &\textbf{9.0} \\ 
{                    } & 5 & 50 & 12.4 &9.6 &\textbf{7.7} \\ 
{                    } & 10 & 100 & 8.9 &7.6 &\textbf{6.4} \\ 
\midrule
{                SVHN} & 2 & 20 & 12.5 &4.9 &\textbf{2.8} \\ 
{                    } & 5 & 50 & 6.2 &4.5 &\textbf{2.3} \\ 
{                    } & 10 & 100 & 4.4 &3.7 &\textbf{1.9} \\ 
\midrule
{            ImageNet} & 2 & 20 & 23.0 &11.5 &\textbf{10.6} \\ 
{                    } & 5 & 50 & 16.0 &10.6 &\textbf{9.0} \\ 
{                    } & 10 & 100 & 11.1 &8.8 &\textbf{7.1} \\ 
\bottomrule
\end{tabular}
}
\label{tab:binwise_acc_estimate}
}
\end{table}

\subsection*{Identifying the least calibrated class}
For identifying the least calibrated classes, in Table~\ref{tab:best-m-ece} we compare the percentage of labeled samples that IPrior and IPrior+TS need to identify the least calibrated top-1 and top-$m$ predicted classes across 5 datasets. 
Table~\ref{tab:best-m-ece} shows that the improvement in efficiency is particularly significant when the classwise calibration performance has large variance across the classes (as shown in Figure~\ref{fig:scatter}), e.g., CIFAR-100, ImageNet and 20 Newsgroups.
\begin{table}[h]
\caption{Percentage of labeled samples needed to identify the least calibrated top-1 and top-$m$ predicted classes.
}
\centering{
\resizebox{\columnwidth}{!}{%
\begin{tabular}{@{}rrrccccc@{}}
\toprule 
& 
& \phantom{a} &  \multicolumn{2}{c}{ECE, Top 1}
& \phantom{a} &  \multicolumn{2}{c}{ECE, Top $m$}\\ 
\cmidrule{4-5} \cmidrule{7-8}
\multicolumn{2}{c}{Dataset} && IPrior &IPrior+TS && IPrior &IPrior+TS \\ \midrule
\multicolumn{2}{c}{     CIFAR-100}  &&88.0 &\textbf{43.0} &&90.0 &\textbf{59.0}\\ 
\multicolumn{2}{c}{      ImageNet}  &&89.6 &\textbf{31.0 }&&90.0 &\textbf{41.2}\\ 
\multicolumn{2}{c}{          SVHN}  &&58.8 &\textbf{40.7} &&88.4 &\textbf{77.6}\\ 
\multicolumn{2}{c}{ 20 Newsgroups}  &&69.0 &\textbf{27.9} &&90.3 &\textbf{50.5}\\ 
\multicolumn{2}{c}{       DBpedia}  &&27.9 & \textbf{8.1} &&89.1 &\textbf{55.6}\\
\bottomrule
\end{tabular}
\label{tab:best-m-ece}
}
}
\end{table}

\subsection*{Comparisons with Alternative Active Learning Algorithms}
\begin{figure*}[h]
    \centering
    \includegraphics[width=\linewidth]{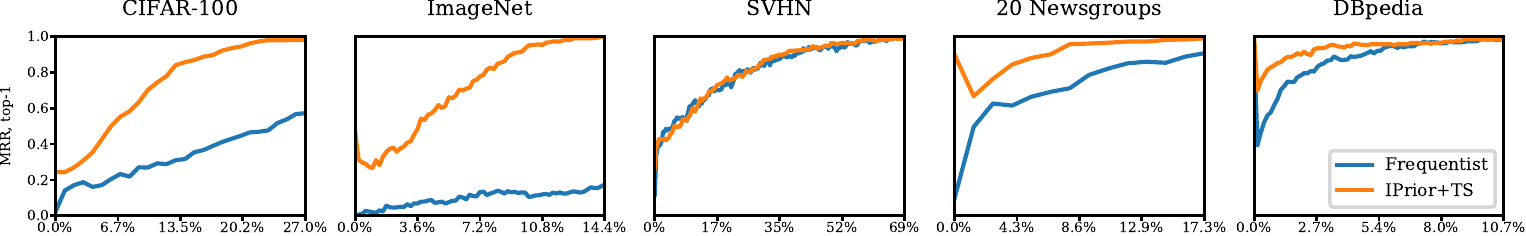}
    \includegraphics[width=\linewidth]{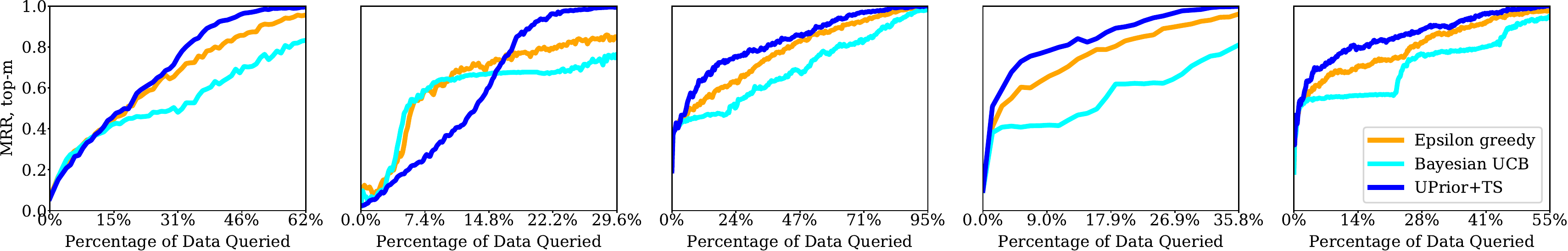}
    \caption{Mean reciprocal rank (MRR) of the classes with the estimated lowest classwise accuracy with the strength of the prior set as 2, comparing Thompson sampling (TS) with epsilon greedy and Bayesian UCB, across five datasets.
    The y-axis is the average MRR over 1000 runs for the percentage of queries, relative to the full test set, as indicated on the x-axis.
    In the upper row $m = 1$, and in the lower row $m = 10$ for CIFAR-100 and ImageNet, and $m = 3$ for the other datasets.}
    \label{fig:alternative_active_learning_methods}
\end{figure*}

There are a variety of other active learning approaches, such as epsilon greedy and Bayesian upper-confidence bound(UCB) methods, that could also be used as alternatives to Thompson sampling. 
\begin{itemize}
    \item Epsilon-greedy: with probability $1-\epsilon$ the arm currently with the greatest expected reward is selected; with probability $\epsilon$ the arm is randomly selected. We set $\epsilon$ as 0.1 in our experiments.
    \item Bayesian upper-confidence bound (UCB): the arm with the greatest upper confidence bound is selected at each step. In our experiments we use the 97.5\% quantile, estimated from 10,000 Monte Carlo samples, as the upper confidence bound.
\end{itemize}
We compare epsilon greedy, Bayesian UCB and Thompson sampling (TS) on the tasks to identify the least accurate and the top-$m$ least accurate predicted classes across five datasets.
Figure~\ref{fig:alternative_active_learning_methods} plots the curves of MRR obtained with three methods as the number of queries increase. 
We use the uninformative prior with prior strength 2 for all three algorithms. 
The results show that the MRR curves of Thompson sampling always converge faster than the curves of epsilon greedy and Bayesian UCB, indicating that Thompson sampling is broadly more reliable and more consistent in terms of efficiency for these tasks.

\subsection*{Comparisons Between IPrior+TS and UPrior+TS}
In this main paper, we compared UPrior, IPrior and IPrior+TS in experimental results, and left out the results of UPrior+TS due to space limits. 
In this subsection, we use the comparison between UPrior+TS and IPrior+TS to demonstrate the influence of informative priors when samples are actively labeled for identifying the least accurate top-1 or top-$m$ predicted classes. We set the strength of both the informative prior and the uninformative prior as 2.

\begin{figure*}[h]
    \centering
    \includegraphics[width=\linewidth]{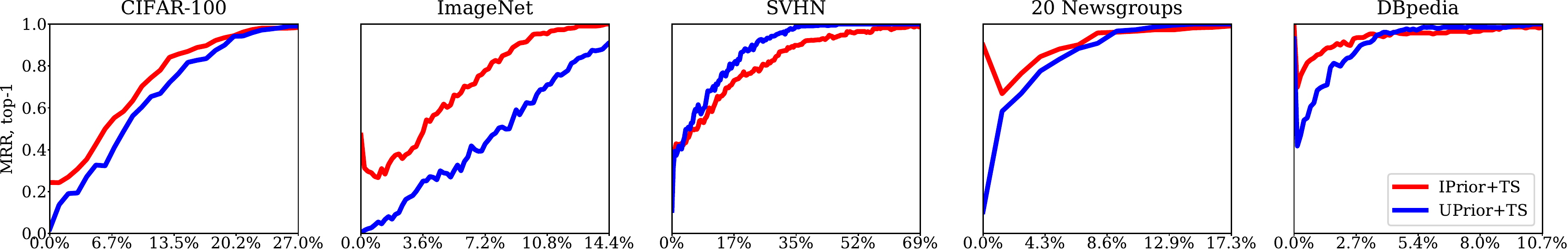}
    \includegraphics[width=\linewidth]{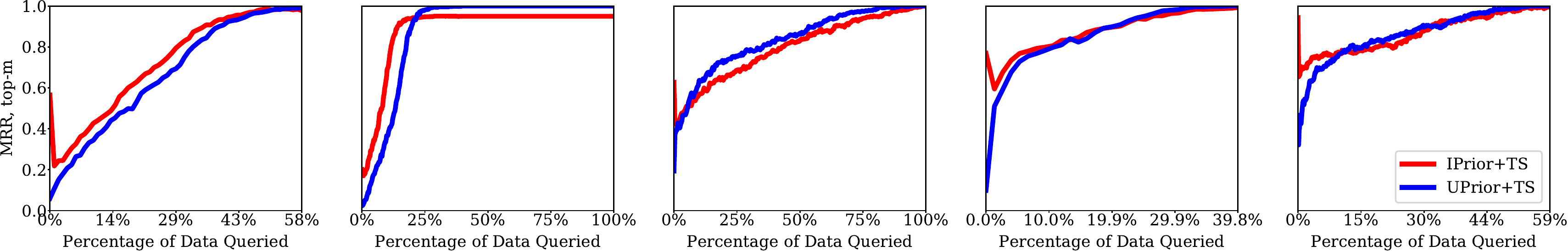}
    \caption{Comparison of the effect of informative (red) and uninformative (blue) priors on identifying the least accurate predicted class with Thompson sampling across 5 datasets.
    The y-axis is the average MRR over 1000 runs for the percentage of queries, relative to the full test set, as indicated on the x-axis.
    In the upper row $m = 1$, and in the lower row $m = 10$ for CIFAR-100 and ImageNet, and $m = 3$ for the other datasets.}
    \label{fig:active_accuracy_informed}
\end{figure*}

The results in Figure~\ref{fig:active_accuracy_informed} illustrate that the informative prior can be helpful when the prior captures the relative ordering of classwise accuracy well (e.g., ImageNet), but less helpful when the difference in classwise accuracy across classes is small and the classwise ordering reflected in the ``self-assessment prior" is more likely to be in error (e.g., SVHN, as shown in Figure~\ref{fig:scatter}.).

In general, across the different estimation tasks, we found that when using active assessment (TS) informative priors (rather than uninformative priors) generally improved performance and rarely hurt it.

\subsection*{Sensitivity Analysis for Hyperparameters}

\begin{figure*}[h]
    \centering 
    \begin{tabular}{c}
    \includegraphics[width=\linewidth]{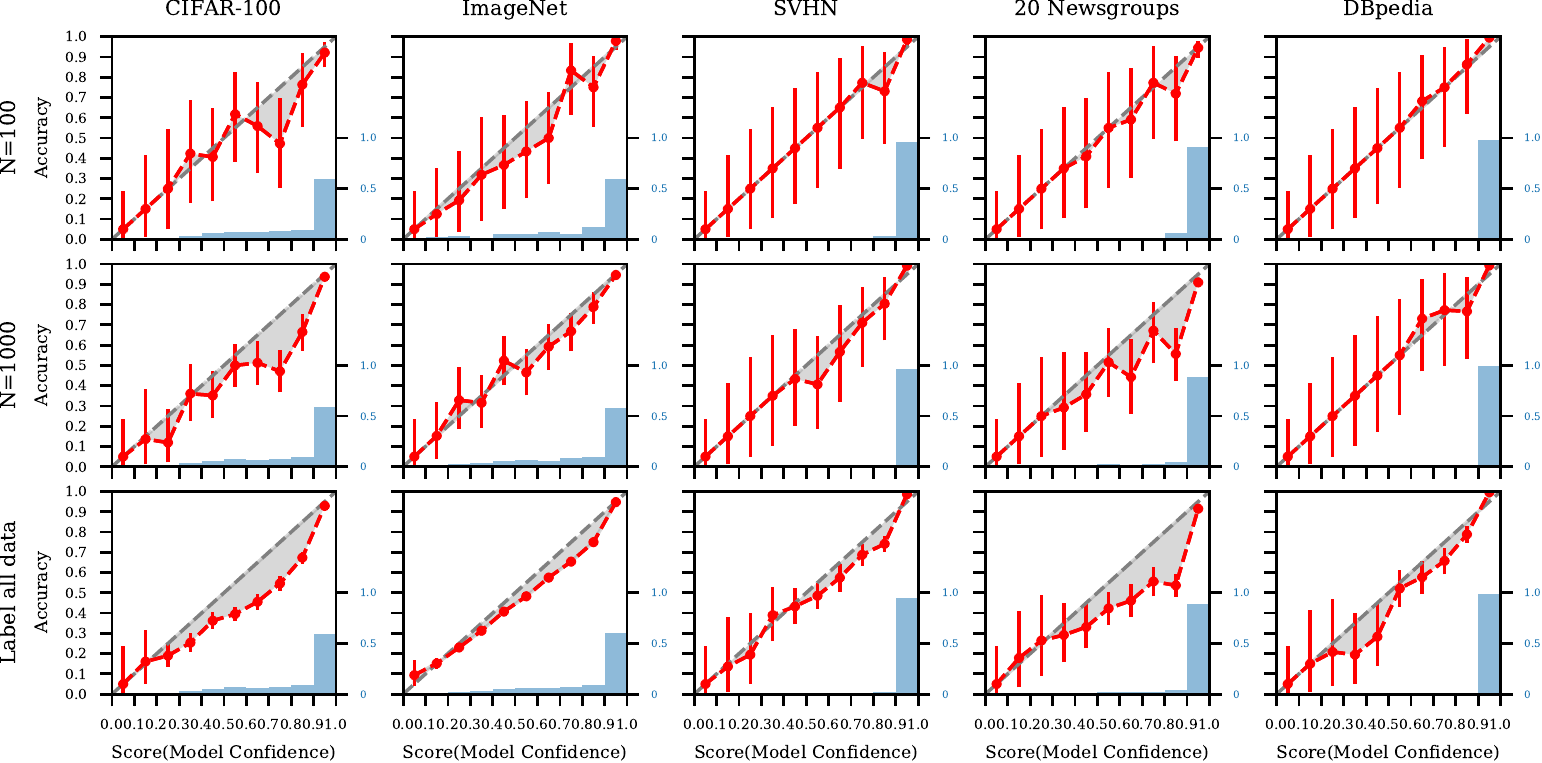}\\
    (a) \\ 
    \\
    \includegraphics[width=\linewidth]{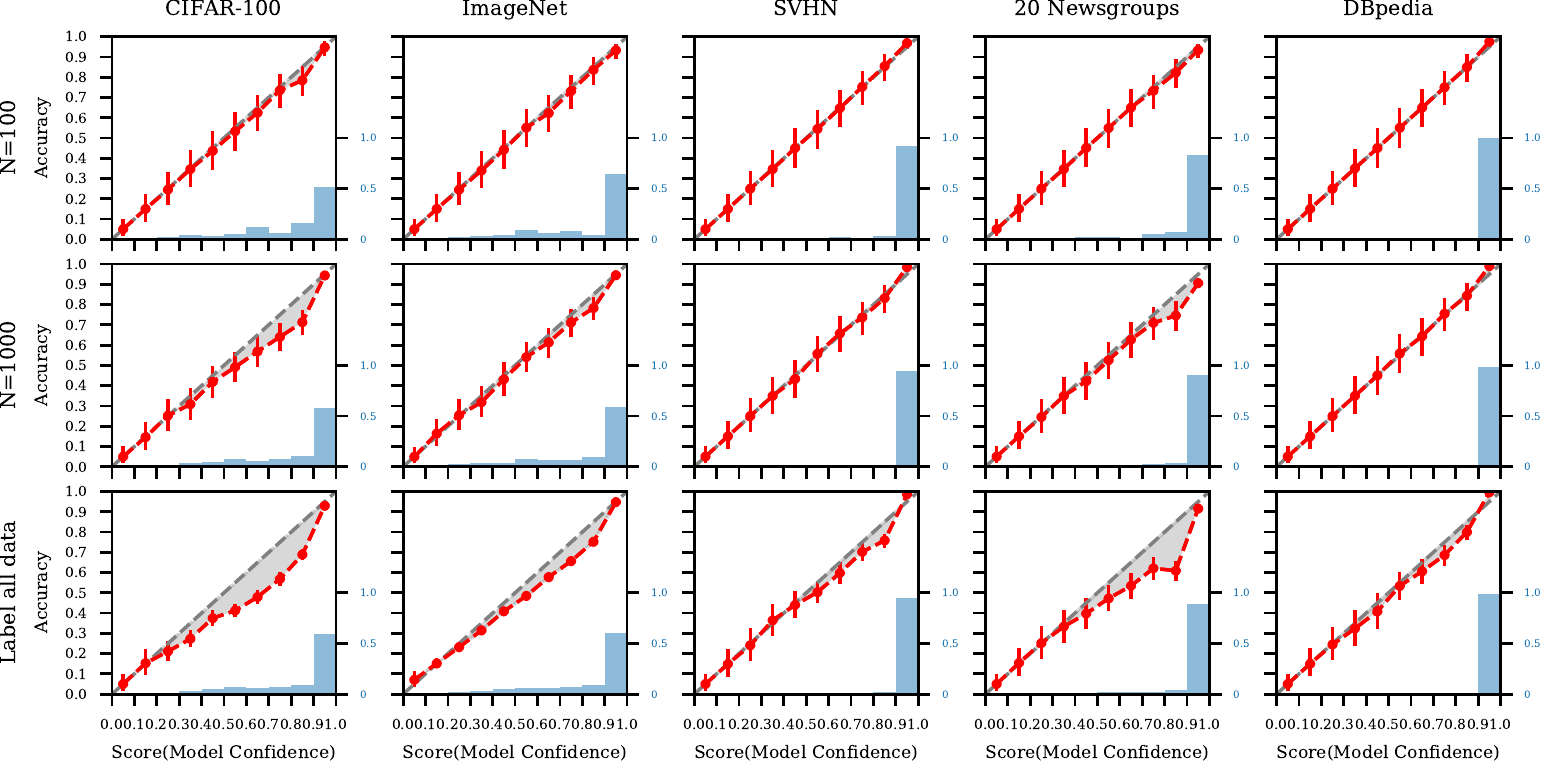}\\
    (b) \\ 
    \end{tabular}
    \caption{
    Bayesian reliability diagrams for five datasets (columns) estimated using varying amounts of  test data (rows) with prior strength ($\alpha_j + \beta_j$ for each bin) set to be (a) 10 and (b) 100 respectively.
    The red circles plot the posterior mean for $\theta_j$ under our Bayesian approach.
    Red bars display 95\% %posterior
    credible intervals.
    Shaded gray areas indicate the estimated magnitudes of the calibration errors, relative to the Bayesian estimates.
    The blue histogram shows the distribution of the scores for $N$ randomly drawn samples.}
    \label{fig:reliability_pseudocount}
\end{figure*}

\begin{figure*}[h]
    \centering 
    \begin{tabular}{c}
    \includegraphics[width=\linewidth]{figures/accuracy_min_mrr_top1_pseudocount2.pdf}\\
    \includegraphics[width=\linewidth]{figures/accuracy_min_mrr_topk_pseudocount2.pdf}\\
    (a) \\ 
    \\
    \includegraphics[width=\linewidth]{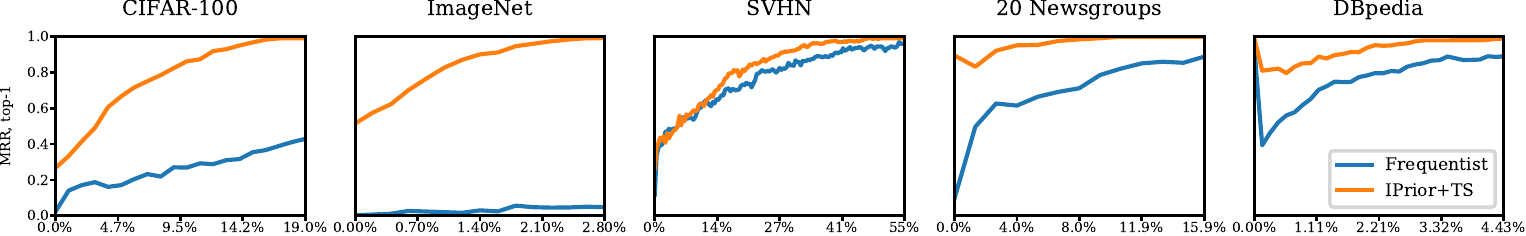}\\
    \includegraphics[width=\linewidth]{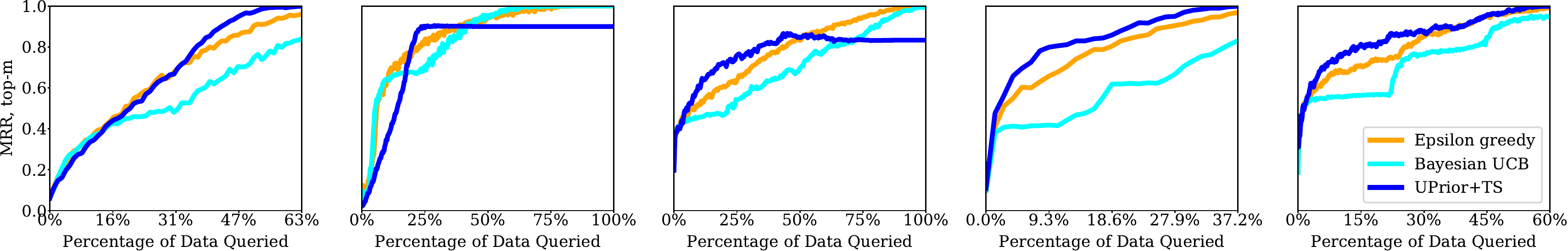}\\
    (b) \\ 
    \\
    \includegraphics[width=\linewidth]{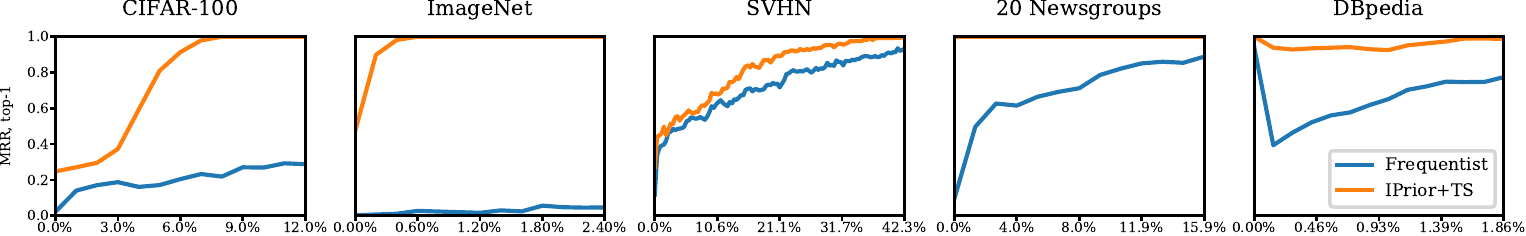}\\
    \includegraphics[width=\linewidth]{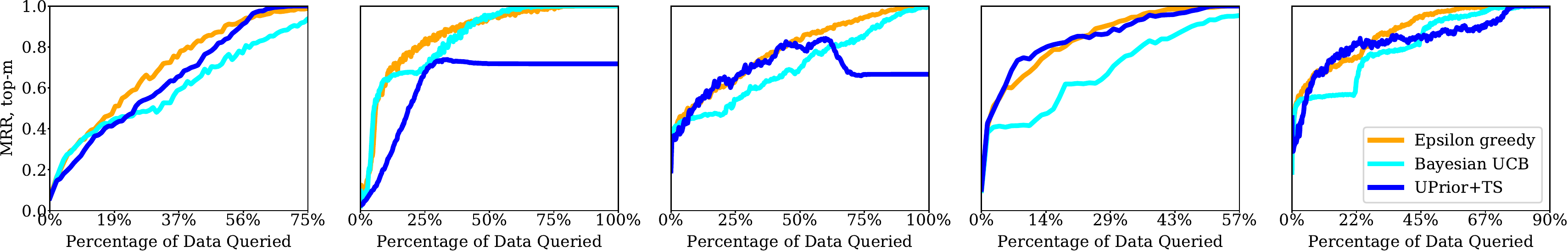}\\
    (c) \\ 
    \end{tabular}
    \caption{
    Mean reciprocal rank (MRR) of the $m$ classes with the estimated lowest classwise accuracy as the strength of the prior varies from (a) 2 to (b) 10 and (c) 100, comparing active learning (with Thompson sampling (IPrior+TS)) with no active learning(Frequentist), across five datasets.
    The y-axis is the average MRR over 1000 runs for the percentage of queries, relative to the full test set, as indicated on the x-axis.
    For each of (a), (b) and (c), in the upper row $m = 1$, and in the lower row $m = 10$ for CIFAR-100 and ImageNet, and $m = 3$ for the other datasets.}
    \label{fig:active_accuracy_pseudocount}
\end{figure*}

\begin{figure*}[ht]
    \centering 
    \begin{tabular}{c}
    \includegraphics[width=\linewidth]{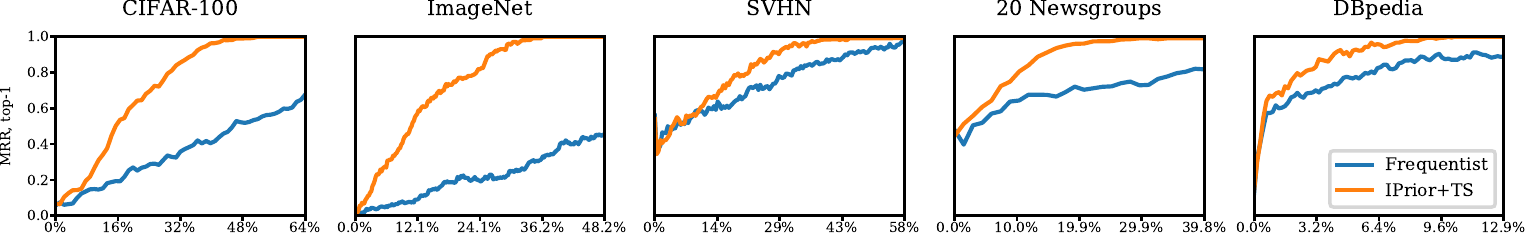}\\
    \includegraphics[width=\linewidth]{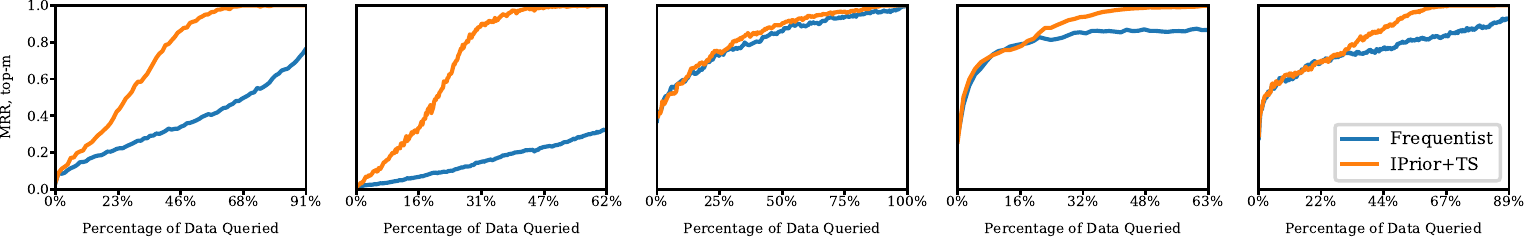}\\
    (a) \\ 
    \\
    \includegraphics[width=\linewidth]{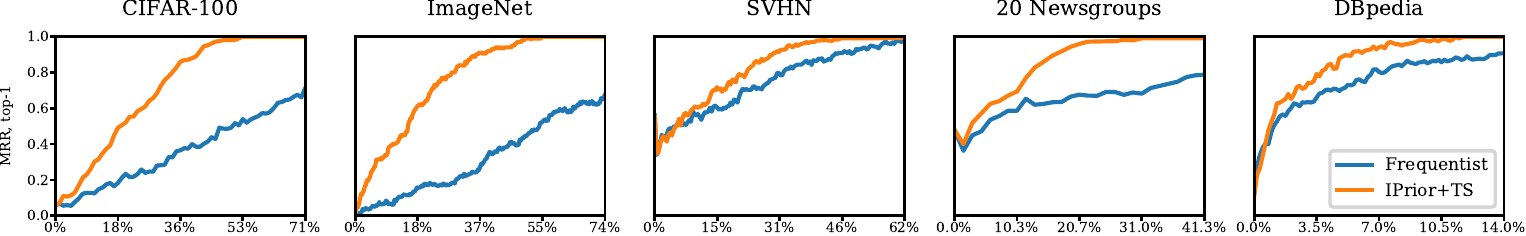}\\
    \includegraphics[width=\linewidth]{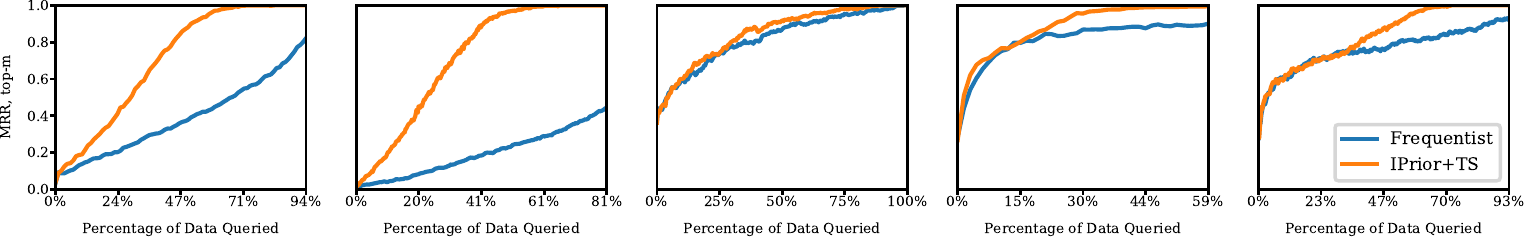}\\
    (b) \\ 
    \\
    \includegraphics[width=\linewidth]{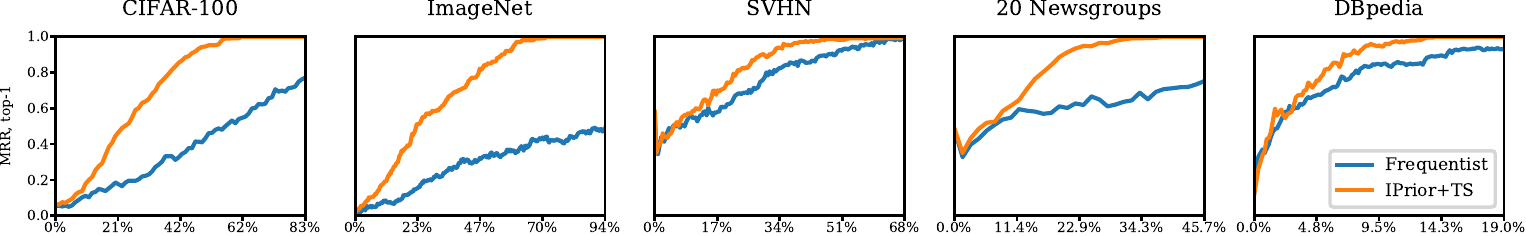}\\
    \includegraphics[width=\linewidth]{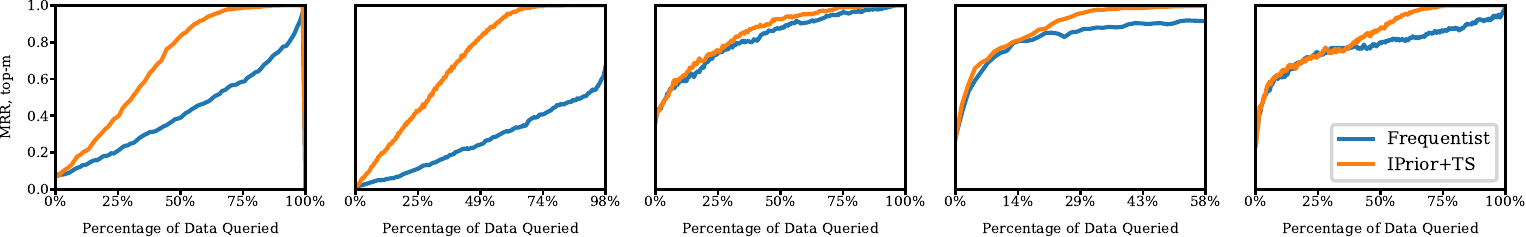}\\
    (c) \\ 
    \end{tabular}
    \caption{
    Mean reciprocal rank (MRR) of the $m$ classes with the estimated highest classwise ECE as the strength of the prior varies from (a) 2 to (b) 5 and (c) 10, comparing active learning (with Thompson sampling (IPrior+TS)) with no active learning(Frequentist), across five datasets.
    The y-axis is the average MRR over 1000 runs for the percentage of queries, relative to the full test set, as indicated on the x-axis.
    For each of (a), (b) and (c), in the upper row $m = 1$, and in the lower row $m = 10$ for CIFAR-100 and ImageNet, and $m = 3$ for the other datasets.}
    \label{fig:active_ece_pseudocount}
\end{figure*}

In Figure~\ref{fig:reliability_pseudocount}, we show Bayesian reliability diagrams for five datasets as the strength of the prior increases from  10 to 100. As the strength of the prior increases, it takes more labeled data to overcome the prior belief that the model is calibrated.
In Figure~\ref{fig:active_accuracy_pseudocount}, we show MRR of the $m$ lowest accurate predicted classes as the strength of the prior increases from 2 to 10 to 100. 
And in Figure~\ref{fig:active_ece_pseudocount}, we show MRR of the $m$ least calibrated  predicted classes as the strength of the prior increase from 2 to 5 and 10. 
From these plots, the proposed approach appears to be relatively robust to the prior strength.

\subsection*{Sensitivity to Cost Matrix Values}
We also investigated the sensitivity of varying the relative cost of mistakes in our cost experiments. 
Results are provided in Table~\ref{tab:active-cost-human} and~\ref{tab:active-cost-superclass}.
We consistently observe that active assessment with an informative prior performs the best, followed by non-active assessment with an informative prior and finally random sampling.

\begin{table}[h]
 \caption{Number of queries required by different methods to achieve a 0.99 mean reciprocal rank(MRR) identifying the class with highest classwise expected cost. A pseudocount of 1 is used in the Dirichlet priors for Bayesian models. The cost type is ``Human.''}
 \label{tab:active-cost-human}
\small
\centering{
\begin{tabular}{rrccc}%c}
\toprule 
Cost   & Top m & UPrior  & IPrior  & IPrior+TS        \\ 
\midrule
1 &1 & 9.6K & 9.4K & 5.0K \\ 
 &10 &10.0K &10.0K & 9.4K \\ 
\midrule
2 &1 & 9.3K & 9.3K & 4.4K \\ 
 &10 & 9.8K &10.0K & 8.4K \\ 
\midrule
5 &1 & 9.5K & 9.7K & 4.5K \\ 
 &10 & 9.6K &10.0K & 7.9K \\ 
\midrule
10 &1 & 9.3K & 9.1K & 2.2K \\ 
 &10 & 9.6K & 9.7K & 7.4K \\ 
\bottomrule
\end{tabular}
}
\end{table}

\begin{table}[h]
\caption{Same setup as Table~\ref{tab:active-cost-human}. The cost type is ``Superclass''.}
\label{tab:active-cost-superclass}
\small
\centering{
\begin{tabular}{rrccc}%c}
\toprule 
Cost   & Top m & UPrior  & IPrior  & IPrior+TS        \\ 
\midrule
1 &1 & 9.9K &10.0K & 2.2K \\ 
 &10 & 9.8K & 9.9K & 5.9K \\ 
\midrule
2 &1 &10.0K &10.0K & 2.2K \\ 
 &10 & 9.9K & 9.9K & 5.2K \\ 
\midrule
5 &1 & 9.9K &10.0K & 1.8K \\ 
 &10 & 9.9K & 9.9K & 5.3K \\ 
\midrule
10 &1 &10.0K & 9.8K & 1.4K \\ 
 &10 & 9.9K & 9.9K & 4.0K \\ 
\bottomrule
\end{tabular}
}
\end{table}

\clearpage
\bibliography{calibration}

\begin{thebibliography}{46}
\providecommand{\natexlab}[1]{#1}
\providecommand{\url}[1]{\texttt{#1}}
\providecommand{\urlprefix}{URL }
\expandafter\ifx\csname urlstyle\endcsname\relax
  \providecommand{\doi}[1]{doi:\discretionary{}{}{}#1}\else
  \providecommand{\doi}{doi:\discretionary{}{}{}\begingroup
  \urlstyle{rm}\Url}\fi

\bibitem[{Aslam, Pavlu, and Yilmaz(2006)}]{aslam2006statistical}
Aslam, J.~A.; Pavlu, V.; and Yilmaz, E. 2006.
\newblock A statistical method for system evaluation using incomplete
  judgments.
\newblock In \emph{Proceedings of the 29th Annual International ACM SIGIR
  Conference on Research and Development in Information Retrieval}, 541--548.

\bibitem[{Benavoli et~al.(2017)Benavoli, Corani, Dem{\v{s}}ar, and
  Zaffalon}]{benavoli2017time}
Benavoli, A.; Corani, G.; Dem{\v{s}}ar, J.; and Zaffalon, M. 2017.
\newblock Time for a change: A tutorial for comparing multiple classifiers
  through {B}ayesian analysis.
\newblock \emph{The Journal of Machine Learning Research} 18(1): 2653--2688.

\bibitem[{DeGroot and Fienberg(1983)}]{degroot1983comparison}
DeGroot, M.~H.; and Fienberg, S.~E. 1983.
\newblock The comparison and evaluation of forecasters.
\newblock \emph{Journal of the Royal Statistical Society: Series D (The
  Statistician)} 32(1-2): 12--22.

\bibitem[{Devlin et~al.(2019)Devlin, Chang, Lee, and
  Toutanova}]{devlin2019bert}
Devlin, J.; Chang, M.-W.; Lee, K.; and Toutanova, K. 2019.
\newblock {BERT}: Pre-training of deep bidirectional transformers for language
  understanding.
\newblock In \emph{NAACL-HLT 2019}, volume~1, 4171--4186.

\bibitem[{Du et~al.(2017)Du, El-Khamy, Lee, and Davis}]{du2017fused}
Du, X.; El-Khamy, M.; Lee, J.; and Davis, L. 2017.
\newblock Fused {DNN}: A deep neural network fusion approach to fast and robust
  pedestrian detection.
\newblock In \emph{Winter Conference on Applications of Computer Vision},
  953--961.

\bibitem[{Freytag, Rodner, and Denzler(2014)}]{freytag2014selecting}
Freytag, A.; Rodner, E.; and Denzler, J. 2014.
\newblock Selecting influential examples: Active learning with expected model
  output changes.
\newblock In \emph{European Conference on Computer Vision}, 562--577. Springer.

\bibitem[{Gal and Ghahramani(2016)}]{gal2016dropout}
Gal, Y.; and Ghahramani, Z. 2016.
\newblock Dropout as a {B}ayesian approximation: Representing model uncertainty
  in deep learning.
\newblock In \emph{International Conference on Machine Learning}, 1050--1059.

\bibitem[{Goutte and Gaussier(2005)}]{goutte2005probabilistic}
Goutte, C.; and Gaussier, E. 2005.
\newblock A probabilistic interpretation of precision, recall and {F}-score,
  with implication for evaluation.
\newblock In \emph{European Conference on Information Retrieval}, 345--359.

\bibitem[{Guo et~al.(2017)Guo, Pleiss, Sun, and
  Weinberger}]{guo2017calibration}
Guo, C.; Pleiss, G.; Sun, Y.; and Weinberger, K.~Q. 2017.
\newblock On calibration of modern neural networks.
\newblock In \emph{International Conference on Machine Learning}, 1321--1330.

\bibitem[{Hardt, Price, and Srebro(2016)}]{hardt2016equality}
Hardt, M.; Price, E.; and Srebro, N. 2016.
\newblock Equality of opportunity in supervised learning.
\newblock In \emph{Advances in neural information processing systems},
  3315--3323.

\bibitem[{He et~al.(2016)He, Zhang, Ren, and Sun}]{he2016deep}
He, K.; Zhang, X.; Ren, S.; and Sun, J. 2016.
\newblock Deep residual learning for image recognition.
\newblock In \emph{Computer Vision and Pattern Recognition}, 770--778.

\bibitem[{Ji, Smyth, and Steyvers(2020)}]{ji2020can}
Ji, D.; Smyth, P.; and Steyvers, M. 2020.
\newblock Can I trust my fairness metric? Assessing fairness with unlabeled
  data and Bayesian inference.
\newblock \emph{Advances in Neural Information Processing Systems} 33.

\bibitem[{Johnson, Jones, and Gardner(2019)}]{johnson2019gold}
Johnson, W.~O.; Jones, G.; and Gardner, I.~A. 2019.
\newblock Gold standards are out and {B}ayes is in: Implementing the cure for
  imperfect reference tests in diagnostic accuracy studies.
\newblock \emph{Preventive Veterinary Medicine} 167: 113--127.

\bibitem[{Kermany et~al.(2018)Kermany, Goldbaum, Cai, Valentim, Liang, Baxter,
  McKeown, Yang, Wu, Yan et~al.}]{kermany2018identifying}
Kermany, D.~S.; Goldbaum, M.; Cai, W.; Valentim, C.~C.; Liang, H.; Baxter,
  S.~L.; McKeown, A.; Yang, G.; Wu, X.; Yan, F.; et~al. 2018.
\newblock Identifying medical diagnoses and treatable diseases by image-based
  deep learning.
\newblock \emph{Cell} 172(5): 1122--1131.

\bibitem[{Komiyama, Honda, and Nakagawa(2015)}]{komiyama2015optimal}
Komiyama, J.; Honda, J.; and Nakagawa, H. 2015.
\newblock Optimal regret analysis of {T}hompson sampling in stochastic
  multi-armed bandit problem with multiple plays.
\newblock In \emph{International Conference on Machine Learning}, 1152--1161.

\bibitem[{Krizhevsky and Hinton(2009)}]{krizhevsky2009learning}
Krizhevsky, A.; and Hinton, G. 2009.
\newblock Learning multiple layers of features from tiny images.
\newblock Technical report, Citeseer.

\bibitem[{Kull, Silva~Filho, and Flach(2017)}]{kull2017beta}
Kull, M.; Silva~Filho, T.; and Flach, P. 2017.
\newblock Beta calibration: A well-founded and easily implemented improvement
  on logistic calibration for binary classifiers.
\newblock In \emph{Artificial Intelligence and Statistics}, 623--631.

\bibitem[{Kumar, Liang, and Ma(2019)}]{kumar2019verified}
Kumar, A.; Liang, P.~S.; and Ma, T. 2019.
\newblock Verified uncertainty calibration.
\newblock In \emph{Advances in Neural Information Processing Systems},
  3787--3798.

\bibitem[{Kumar and Raj(2018)}]{kumar2018classifier}
Kumar, A.; and Raj, B. 2018.
\newblock Classifier risk estimation under limited labeling resources.
\newblock In \emph{Pacific-Asia Conference on Knowledge Discovery and Data
  Mining}, 3--15. Springer.

\bibitem[{Lakshminarayanan, Pritzel, and
  Blundell(2017)}]{lakshminarayanan2017simple}
Lakshminarayanan, B.; Pritzel, A.; and Blundell, C. 2017.
\newblock Simple and scalable predictive uncertainty estimation using deep
  ensembles.
\newblock In \emph{Advances in Neural Information Processing Systems},
  6402--6413.

\bibitem[{Lang(1995)}]{lang1995newsweeder}
Lang, K. 1995.
\newblock NewsWeeder: Learning to filter netnews.
\newblock In \emph{Proceedings of the Twelfth International Conference on
  International Conference on Machine Learning}, 331--339.

\bibitem[{Li and Kanoulas(2017)}]{li2017active}
Li, D.; and Kanoulas, E. 2017.
\newblock Active sampling for large-scale information retrieval evaluation.
\newblock In \emph{Proceedings of the 2017 ACM on Conference on Information and
  Knowledge Management}, 49--58.

\bibitem[{Marshall and Spiegelhalter(1998)}]{marshall1998league}
Marshall, E.~C.; and Spiegelhalter, D.~J. 1998.
\newblock League tables of in vitro fertilisation clinics: How confident can we
  be about the rankings.
\newblock \emph{British Medical Journal} 316: 1701--1704.

\bibitem[{Moffat, Webber, and Zobel(2007)}]{moffat2007strategic}
Moffat, A.; Webber, W.; and Zobel, J. 2007.
\newblock Strategic system comparisons via targeted relevance judgments.
\newblock In \emph{Proceedings of the 30th Annual International ACM SIGIR
  Conference on Research and Development in Information Retrieval}, 375--382.

\bibitem[{Netzer et~al.(2011)Netzer, Wang, Coates, Bissacco, Wu, and
  Ng}]{netzer2011reading}
Netzer, Y.; Wang, T.; Coates, A.; Bissacco, A.; Wu, B.; and Ng, A.~Y. 2011.
\newblock Reading digits in natural images with unsupervised feature learning.
\newblock \emph{NIPS Workshop on Deep Learning and Unsupervised Feature
  Learning} .

\bibitem[{Nguyen, Ramanan, and Fowlkes(2018)}]{nguyen2018active}
Nguyen, P.; Ramanan, D.; and Fowlkes, C. 2018.
\newblock Active testing: An efficient and robust framework for estimating
  accuracy.
\newblock In \emph{International Conference on Machine Learning}, 3759--3768.

\bibitem[{Niculescu-Mizil and Caruana(2005)}]{niculescu2005predicting}
Niculescu-Mizil, A.; and Caruana, R. 2005.
\newblock Predicting good probabilities with supervised learning.
\newblock In \emph{International Conference on Machine Learning}, 625--632.

\bibitem[{Nixon et~al.(2019)Nixon, Dusenberry, Zhang, Jerfel, and
  Tran}]{Nixon_2019_CVPR_Workshops}
Nixon, J.; Dusenberry, M.~W.; Zhang, L.; Jerfel, G.; and Tran, D. 2019.
\newblock Measuring Calibration in Deep Learning.
\newblock In \emph{The IEEE Conference on Computer Vision and Pattern
  Recognition Workshops}.

\bibitem[{Ovadia et~al.(2019)Ovadia, Fertig, Ren, Nado, Sculley, Nowozin,
  Dillon, Lakshminarayanan, and Snoek}]{snoek2019can}
Ovadia, Y.; Fertig, E.; Ren, J.; Nado, Z.; Sculley, D.; Nowozin, S.; Dillon,
  J.~V.; Lakshminarayanan, B.; and Snoek, J. 2019.
\newblock Can you trust your model's uncertainty? Evaluating predictive
  uncertainty under dataset shift.
\newblock In \emph{Advances in Neural Information Processing Systems},
  13969--13980.

\bibitem[{Rahman et~al.(2018)Rahman, Kutlu, Elsayed, and
  Lease}]{rahman2018efficient}
Rahman, M.~M.; Kutlu, M.; Elsayed, T.; and Lease, M. 2018.
\newblock Efficient test collection construction via active learning.
\newblock \emph{arXiv preprint arXiv:1801.05605} .

\bibitem[{Rahman, Kutlu, and Lease(2019)}]{rahman2019constructing}
Rahman, M.~M.; Kutlu, M.; and Lease, M. 2019.
\newblock Constructing test collections using multi-armed bandits and active
  learning.
\newblock In \emph{The World Wide Web Conference}, 3158--3164.

\bibitem[{Russakovsky et~al.(2015)Russakovsky, Deng, Su, Krause, Satheesh, Ma,
  Huang, Karpathy, Khosla, Bernstein et~al.}]{russakovsky2015imagenet}
Russakovsky, O.; Deng, J.; Su, H.; Krause, J.; Satheesh, S.; Ma, S.; Huang, Z.;
  Karpathy, A.; Khosla, A.; Bernstein, M.; et~al. 2015.
\newblock Imagenet large scale visual recognition challenge.
\newblock \emph{International Journal of Computer Vision} 115(3): 211--252.

\bibitem[{Russo(2016)}]{russo2016simple}
Russo, D. 2016.
\newblock Simple {B}ayesian algorithms for best arm identification.
\newblock In \emph{Conference on Learning Theory}, 1417--1418.

\bibitem[{Russo et~al.(2018)Russo, Van~Roy, Kazerouni, Osband, Wen
  et~al.}]{russo2018tutorial}
Russo, D.~J.; Van~Roy, B.; Kazerouni, A.; Osband, I.; Wen, Z.; et~al. 2018.
\newblock A tutorial on Thompson sampling.
\newblock \emph{Foundations and Trends in Machine Learning} 11(1): 1--96.

\bibitem[{Sabharwal and Sedghi(2017)}]{sabharwal2017good}
Sabharwal, A.; and Sedghi, H. 2017.
\newblock How good are my predictions? Efficiently approximating
  precision-recall curves for massive datasets.
\newblock In \emph{Conference on Uncertainty in Artificial Intelligence}.

\bibitem[{Sanyal et~al.(2018)Sanyal, Kusner, Gasc{\'o}n, and
  Kanade}]{sanyal2018tapas}
Sanyal, A.; Kusner, M.~J.; Gasc{\'o}n, A.; and Kanade, V. 2018.
\newblock {TAPAS}: Tricks to accelerate (encrypted) prediction as a service.
\newblock In \emph{International Conference on Machine Learning}, volume~80,
  4490--4499. PMLR.

\bibitem[{Sawade et~al.(2010)Sawade, Landwehr, Bickel, and
  Scheffer}]{sawade2010active}
Sawade, C.; Landwehr, N.; Bickel, S.; and Scheffer, T. 2010.
\newblock Active risk estimation.
\newblock In \emph{International Conference on Machine Learning}, 951--958.

\bibitem[{Settles(2012)}]{settles2012active}
Settles, B. 2012.
\newblock \emph{Active Learning}.
\newblock Synthesis Lectures on AI and ML. Morgan Claypool.

\bibitem[{Thompson(1933)}]{thompson1933likelihood}
Thompson, W.~R. 1933.
\newblock On the likelihood that one unknown probability exceeds another in
  view of the evidence of two samples.
\newblock \emph{Biometrika} 25(3/4): 285--294.

\bibitem[{Vezhnevets, Buhmann, and Ferrari(2012)}]{vezhnevets2012active}
Vezhnevets, A.; Buhmann, J.~M.; and Ferrari, V. 2012.
\newblock Active learning for semantic segmentation with expected change.
\newblock In \emph{2012 IEEE Conference on Computer Vision and Pattern
  Recognition}, 3162--3169. IEEE.

\bibitem[{Voorhees(2018)}]{voorhees2018building}
Voorhees, E.~M. 2018.
\newblock On building fair and reusable test collections using bandit
  techniques.
\newblock In \emph{Proceedings of the 27th ACM International Conference on
  Information and Knowledge Management}, 407--416.

\bibitem[{Welinder, Welling, and Perona(2013)}]{welinder2013lazy}
Welinder, P.; Welling, M.; and Perona, P. 2013.
\newblock A lazy man's approach to benchmarking: Semisupervised classifier
  evaluation and recalibration.
\newblock In \emph{Proceedings of the IEEE Conference on Computer Vision and
  Pattern Recognition}, 3262--3269.

\bibitem[{Yao et~al.(2017)Yao, Xiao, Wang, Viswanath, Zheng, and
  Zhao}]{yao2017complexity}
Yao, Y.; Xiao, Z.; Wang, B.; Viswanath, B.; Zheng, H.; and Zhao, B.~Y. 2017.
\newblock Complexity vs. performance: Empirical analysis of machine learning as
  a service.
\newblock In \emph{Internet Measurement Conference}, 384--397.

\bibitem[{Yilmaz and Aslam(2006)}]{yilmaz2006estimating}
Yilmaz, E.; and Aslam, J.~A. 2006.
\newblock Estimating average precision with incomplete and imperfect judgments.
\newblock In \emph{Proceedings of the 15th ACM International Conference on
  Information and Knowledge Management}, 102--111.

\bibitem[{Zadrozny and Elkan(2002)}]{zadrozny2002transforming}
Zadrozny, B.; and Elkan, C. 2002.
\newblock Transforming classifier scores into accurate multiclass probability
  estimates.
\newblock In \emph{International Conference on Knowledge Discovery and Data
  Mining}, 694--699.

\bibitem[{Zhang, Zhao, and LeCun(2015)}]{zhang2015character}
Zhang, X.; Zhao, J.; and LeCun, Y. 2015.
\newblock Character-level convolutional networks for text classification.
\newblock In \emph{Advances in Neural Information Processing Systems},
  649--657.

\end{thebibliography}

\end{document}